\documentclass[preprint,12pt,authoryear]{elsarticle}

\usepackage{amssymb}
\usepackage{amsmath}

\usepackage{lineno}

\journal{Neural Networks}

\begin{document}

\begin{frontmatter}



\title{Oscillations enhance time-series prediction in reservoir computing with feedback} 


\author[label1]{Yuji Kawai} 
\author[label2]{Takashi Morita} 
\author[label3,label1]{Jihoon Park} 
\author[label4,label1,label2,label3]{Minoru Asada} 

\affiliation[label1]{organization={Symbiotic Intelligent Systems Research Center, Institute for Open and Transdisciplinary Research Initiatives, Osaka University},
            addressline={1-1 Yamadaoka},
            city={Suita},
            postcode={565--0871},
            state={Osaka},
            country={Japan}}

\affiliation[label2]{organization={Academy of Emerging Sciences, Chubu University},
            addressline={1200 Matsumoto-cho},
            city={Kasugai},
            postcode={487--8501},
            state={Aichi},
            country={Japan}}

\affiliation[label3]{organization={Center for Information and Neural Networks, National Institute of Information and Communications Technology},
            addressline={1-4 Yamadaoka},
            city={Suita},
            postcode={565--0871},
            state={Osaka},
            country={Japan}}

\affiliation[label4]{organization={International Professional University of Technology in Osaka},
            addressline={3-3-1 Umeda},
            city={Kita-ku},
            postcode={530--0001},
            state={Osaka},
            country={Japan}}

\begin{abstract}
Reservoir computing, a machine learning framework used for modeling the brain, can predict temporal data with little observations and minimal computational resources.
However, it is difficult to accurately reproduce the long-term target time series because the reservoir system becomes unstable.
This predictive capability is required for a wide variety of time-series processing, including predictions of motor timing and chaotic dynamical systems.
This study proposes oscillation-driven reservoir computing (ODRC) with feedback, where oscillatory signals are fed into a reservoir network to stabilize the network activity and induce complex reservoir dynamics.
The ODRC can reproduce long-term target time series more accurately than conventional reservoir computing methods in a motor timing and chaotic time-series prediction tasks.
Furthermore, it generates a time series similar to the target in the unexperienced period, that is, it can learn the abstract generative rules from limited observations.
Given these significant improvements made by the simple and computationally inexpensive implementation, the ODRC would serve as a practical model of various time series data.
Moreover, we will discuss biological implications of the ODRC, considering it as a model of neural oscillations and their cerebellar processors.
\end{abstract}








\end{frontmatter}



\section{Introduction}
Oscillations with a wide range of frequencies are a prominent feature of brain activity and are associated with various functions such as perception~\citep{bacsar2000brain, hipp2011oscillatory}, memory~\citep{bacsar2000brain, duzel2010brain, buzsaki2013memory}, and learning~\citep{caplan2001distinct, seager2002oscillatory}.
For example, trace classical conditioning of the eyeblink response, which is one of the best-understood models of timing learning, relies on the cerebellum and hippocampal theta (3--7 Hz) oscillations~\citep{solomon1986hippocampus, nokia2009hippocampal, hoffmann2009cerebellar, berry2011hippocampal}.
The cerebellum is the main region responsible for eyeblink conditioning; however, if a stimulus-free interval exists between stimuli, i.e., trace conditioning, cerebellar activity evoked by the first stimulus must be sustained during the trace interval.
It has been hypothesized that persistent theta oscillations from the hippocampus drive the cerebellum, enabling the cerebellar learning for trace conditioning~\citep{solomon1986hippocampus, clark2002classical, kalmbach2009interactions, hoffmann2009cerebellar, weiss2015impact}.
However, the general mechanisms by which these oscillations contribute to learning remain unclear.

To model temporal/sequential processing, including timing learning, reservoir computing~\citep{jaeger2001echo, jaeger2004harnessing}, a type of recurrent neural networks, has been utilized~\citep{laje2013robust, kawai2023learning}.
Spontaneous, self-sustained reservoir activity is required to accomplish a timing task with a stimulus-free interval.
However, this activity is often associated with orbital instability, in which slight differences in the initial values and noise increase exponentially over time~\citep{sompolinsky1988chaos, van1996chaos, london2010sensitivity}.
Standard reservoir computing comprises a fixed randomly connected reservoir network that creates complex dynamics, and its network states are integrated by a linear readout to obtain a desired time series.
Accordingly, previous studies have explored alternative learning algorithms and/or network architectures to improve the stability.
\cite{laje2013robust} proposed a reservoir computing model, called  innate training, to learn motor timing, where connection weights in a reservoir were modulated to stabilize and denoise the reservoir activity.
As a model beyond its performance, \cite{kawai2023learning, kawai2023reservoir} proposed a reservoir of basal dynamics (reBASICS) to accomplish the timing task with a long-term interval.
reBASICS comprises several small reservoirs, whose small network size suppresses the orbital instability.
Although reBASICS reproduces a learned time series, it cannot generalize it; that is, it cannot generate a time series analogous to the learned one during the unlearned period.
For example, reBASICS can generate a time series with a peak at a given time interval; however, after that interval, it generates cluttered outputs.
By contrast, the brain has excellent generalization capabilities and can learn the rules behind limited observations to generate analogous outputs.

The key to the generalizability in reservoir computing is the output feedback (i.e., autoregression).
Feeding back the readout outputs to the reservoir as inputs allows it to learn to predict the target time series at the current time from the outputs from one time ago~\citep{jaeger2004harnessing, kim2021teaching, sussillo2009generating, vlachas2020backpropagation}.
That is, the reservoir learns the differences from the previous outputs, which reduces the learning load.
This one-time-ahead prediction and feedback are repeated to generate the target time series.
However, when such systems learn a complex time series, the system becomes unstable, and the outputs and targets become misaligned during long-term prediction~\citep{pathak2017using, sussillo2009generating, vlachas2020backpropagation, platt2022systematic}.
Therefore, an additional method to reduce the instability of reservoir systems is required for long-term prediction.

We hypothesized that oscillation inputs to a reservoir with feedback can stabilize reservoir activity, enabling the reservoir to learn long-term timing and generalize the target time series.
Time-varying stable inputs suppress the instability of random recurrent networks~\citep{rajan2010stimulus, takasu2024suppression}.
\cite{vincent2016driving, vincent2020learning} proposed the provision of oscillator (sinusoidal) inputs to reservoirs.
Multiple sinusoidal inputs with different frequencies have been shown to reduce reservoir instability and drive complex reservoir dynamics, allowing timing learning~\citep{vincent2016driving}.
However, this reservoir system is not equipped with output feedback and therefore lacks generalizability~\citep{kawai2024oscillator}.
In this study, we provided oscillation inputs into the reservoir with feedback (hereafter, referred to as oscillation-driven reservoir computing, ODRC) and evaluated its performance in motor timing and chaotic time-series prediction tasks.
We show that the ODRC can learn motor timing with a very long interval, is robust to noise, and can predict chaotic time series accurately and generalize them.
Finally, we discuss how our system can be the model for learning in the brain.

\section{Results}
\subsection{Oscillation-driven reservoir computing}
\begin{figure}[tb]
	\begin{center}
		\includegraphics*[width=13.0cm]{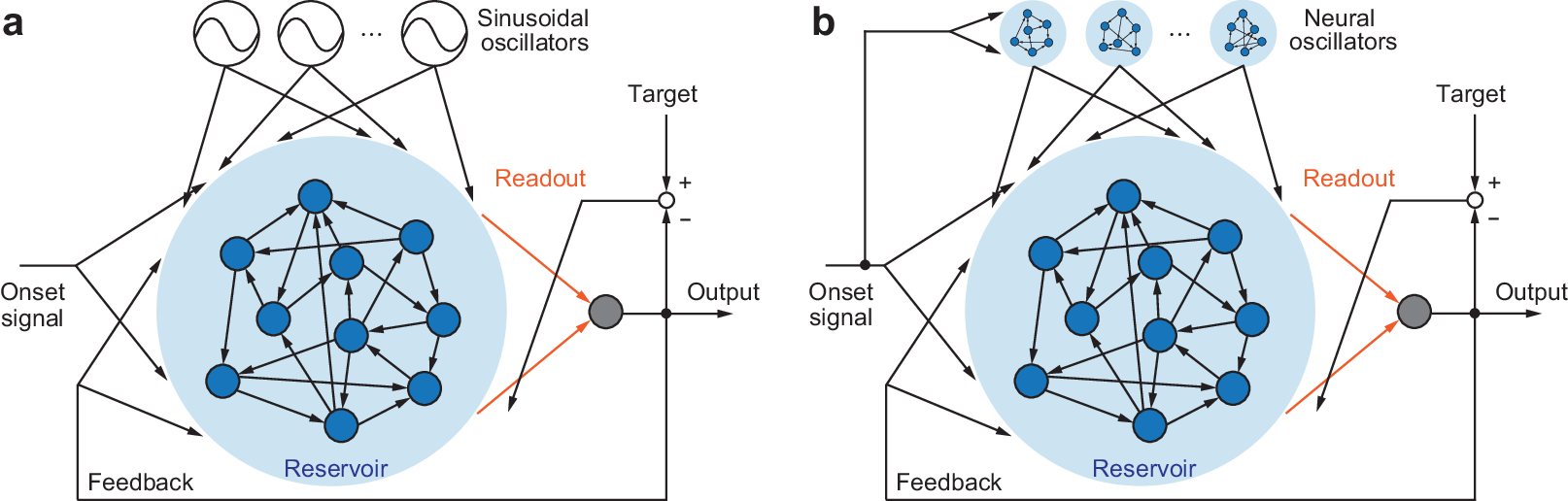}
		\caption{Overview of architectures of oscillation-driven reservoir computing with feedback (ODRC).
			In both cases, oscillators drive a recurrent neural network (reservoir) with fixed random connectivity.
			The reservoir activity is integrated through readouts (orange arrows) of linear summation to produce outputs.
			The readout outputs are fed back into the reservoir.
			Only the readout weights are trained with recursive least squares, whereas other weights are fixed.
			{\bf a} In sine ODRC, the oscillators are sinusoidal.
			Their frequencies and initial phases are randomly determined.
			{\bf b} In neural ODRC, the oscillators consist of random neural networks.
			Reducing their network size makes their network acitivity oscillatory.
			Those oscillations have different frequencies.}
		\label{fig:model}
	\end{center}
\end{figure}

The sine ODRC obtains inputs from multiple sinusoidal oscillators (Fig.~\ref{fig:model}a).
The oscillators comprise different random frequencies in a given range and random initial phases, which are consistent among the trials.
The reservoir receives an additional input; that is, an onset signal.
This is a scalar single pulse at time zero to reduce the initial value dependence of the reservoir activity, that is, it serves as a phase resetting~\citep{laje2013robust, kawai2023learning}.
The reservoir is a randomly connected network comprising firing-rate neural units with a time constant of 10 ms.

Their initial states are randomly determined.
Each time, the output is obtained by the linear summation (readout) of the reservoir states.
The output is then fed back into the reservoir at the subsequent time.
The connection weights for the oscillator inputs, onset signal input, reservoir, and feedback are fixed through learning, whereas the readout weights are modulated to minimize errors between the outputs and target time series.
The readout can be trained using recursive least squares, which is an online learning method~\citep{haykin2009neural}.

Nonlinear neural oscillators, other than pure sine oscillations, can be used for the ODRC.
In this study, limit-cycle oscillations created by small random neural networks~\citep{doyon1994bifurcations, kawai2023learning} are utilized as neural oscillators (Fig.~\ref{fig:model}b).
These neural oscillators have different random connectivity for each oscillator, allowing them to generate stable limit cycles with different frequencies.
In each neural oscillator, activity of a randomly chosen neural unit is regarded as an output of the oscillator and is input to the reservoir.

\subsection{Motor timing}
In the motor timing task, the target is a one-dimensional time series with a single Gaussian peak of amplitude 1.0 and a standard deviation 30 ms at a given time interval and a constant of 0.2, except during its pulse period.
The interval was set from 1 s to 120 s.
The task period was set to the interval $+$ 150 ms.
The square of the correlation coefficient ($R^2$) between the output of the model in an untrained test trial and the target time series was used to evaluate the performance.

This aforementioned task cannot be accomplished by standard reservoir computing (echo state networks and FORCE learning~\citep{sussillo2009generating}) without oscillations~\citep{laje2013robust, kawai2023spatiotemporal}.
The lack of inputs during this interval necessitates the reservoir to continue generating self-sustained activity.
The activity must satisfy orbital stability; that is, the trajectories must not vary from trial to trial, even with different initial states.

\subsubsection{Sine ODRC}
\begin{figure}[!tbhp]
	\begin{center}
		\includegraphics*[width=13cm]{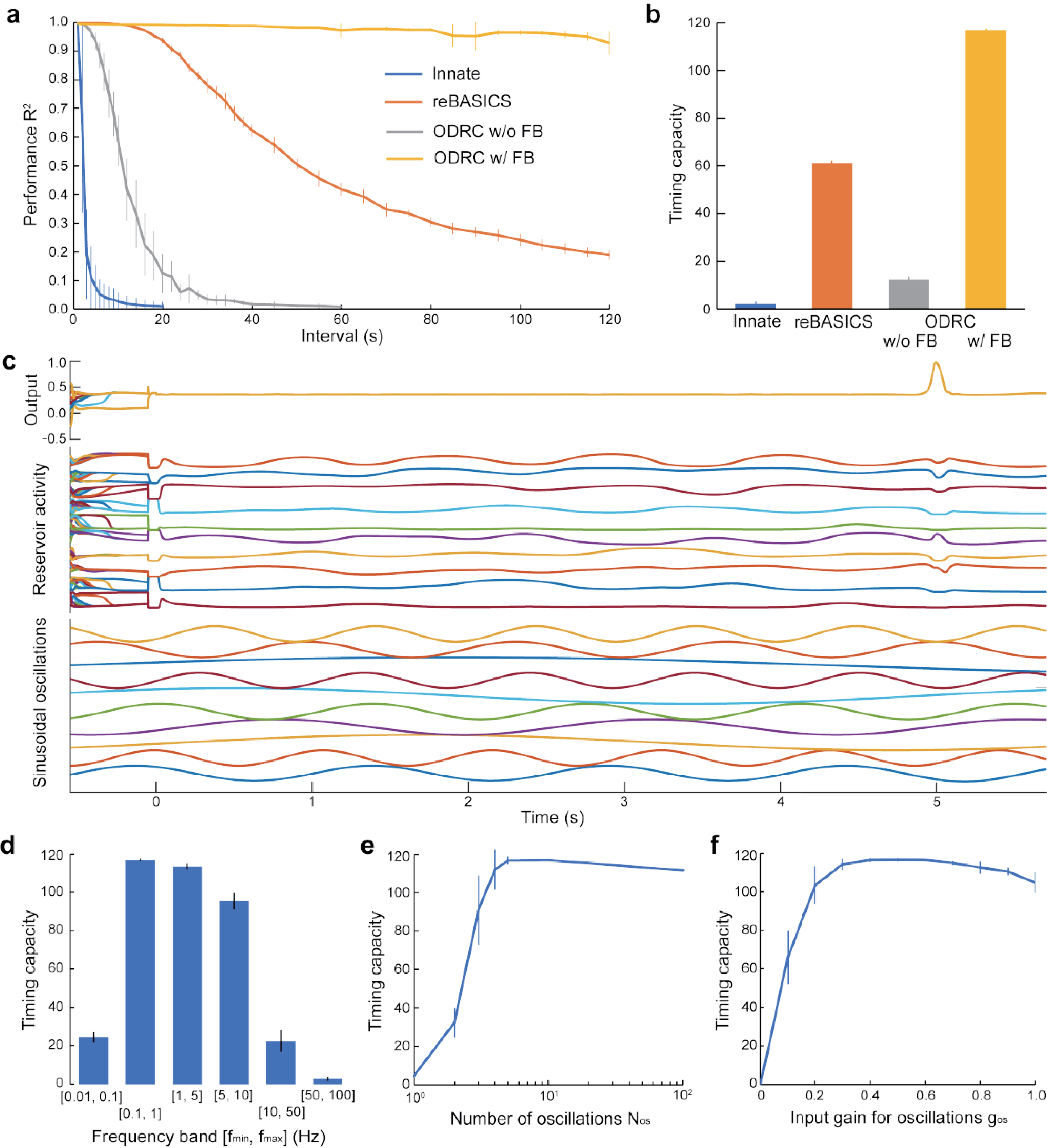}
		\caption{Results of sinusoidal oscillator-driven reservoir computing (ODRC) for the motor timing task.
			The performance was averaged over ten networks (mean $\pm$ s.d.).
			{\bf a} Performance $R^2$ for up to 120 s of intervals, which is defined as the square of the correlation coefficient between the target time series and model outputs.
			The performance of the ODRC with feedback (w/ FB), ODRC without feedback (w/o FB)~\citep{vincent2016driving}, innate training method~\citep{laje2013robust}, and reservoir of basal dynamics (reBASICS)~\citep{kawai2023learning} was evaluated.
			{\bf b} Performance curves summarized as the areas under the curves, referred to as timing capacities.
			{\bf c} Trajectories of the output (top), reservoir activity of ten neural units (middle), and oscillations (bottom) in the ODRC w/ FB in the 5-s task.
			These are the overlaid curves from ten test trials.
			{\bf d}, {\bf e}, {\bf f} Timing capacities of the ODRC w/ FB when varying the frequency band of the oscillators, the number of oscillators, and the gain of oscillation inputs, respectively.}
		\label{fig:sine_timing}
	\end{center}
\end{figure}

We evaluated the performances of the sine ODRC with feedback, without feedback~\citep{vincent2016driving}, innate training~\citep{laje2013robust}, and reBASICS~\citep{kawai2023learning} (Fig.~\ref{fig:sine_timing}a).
We used ten sinusoidal oscillators for the ODRC, whose frequency band was $[0.1, 1]$ Hz.
The $R^2$ of the ODRC with feedback was greater than 0.9, even at an interval of 120 s, showing the best performance among all the methods.
Compared to the time constant of the neural units (10 ms), the ODRC with feedback was capable of representing long intervals.
Fig.~\ref{fig:sine_timing}b shows timing capacities, as defined by the area under the performance $R^2$ curves up to 120s.
The performance of the ODRC with feedback was significantly better than that of the ODRC without feedback, innate method, and reBASICS.

Fig.~\ref{fig:sine_timing}c shows examples of the readout output (top), reservoir activity (middle), and oscillator inputs (bottom) of the ODRC with feedback at an interval of 5 s.
Generally, if the gain in reservoir weights is sufficiently large to self-sustain the reservoir activity, the activity becomes chaotic.
However, the oscillator inputs suppressed the orbital instability associated with chaos, resulting in little variability in reservoir activity among trials.
Using such a stable complex reservoir activity, the desired readout output was correctly obtained without variability.

We examined the parameter sensitivity of the ODRC using feedback.
The performance depends on the frequency band of the oscillators (Fig.~\ref{fig:sine_timing}d).
It peaked at $[0.1, 1]$ Hz, with lower or higher frequencies resulting in lower performance.
Subsequently, when the number of oscillators was varied, the performance was saturated at more than five oscillators (Fig.~\ref{fig:sine_timing}e).
These results indicate that the small- and low-frequency oscillator inputs drive the complex dynamics of reservoirs.
The gain of the oscillator input weights exhibited a medium optimum value (Fig.~\ref{fig:sine_timing}f).
If the gain was excessively large, the dynamics of the reservoir were completely entrained in the oscillator inputs, resulting in a lack of dynamic complexity.
The gain of the reservoir's recurrent connection weights is also an important parameter for reservoir computing, possessing an optimal value greater than 1 (\ref{appendix1}).

\subsubsection{Neural ODRC}
\begin{figure}[!tb]
	\begin{center}
		\includegraphics*[width=13.0cm]{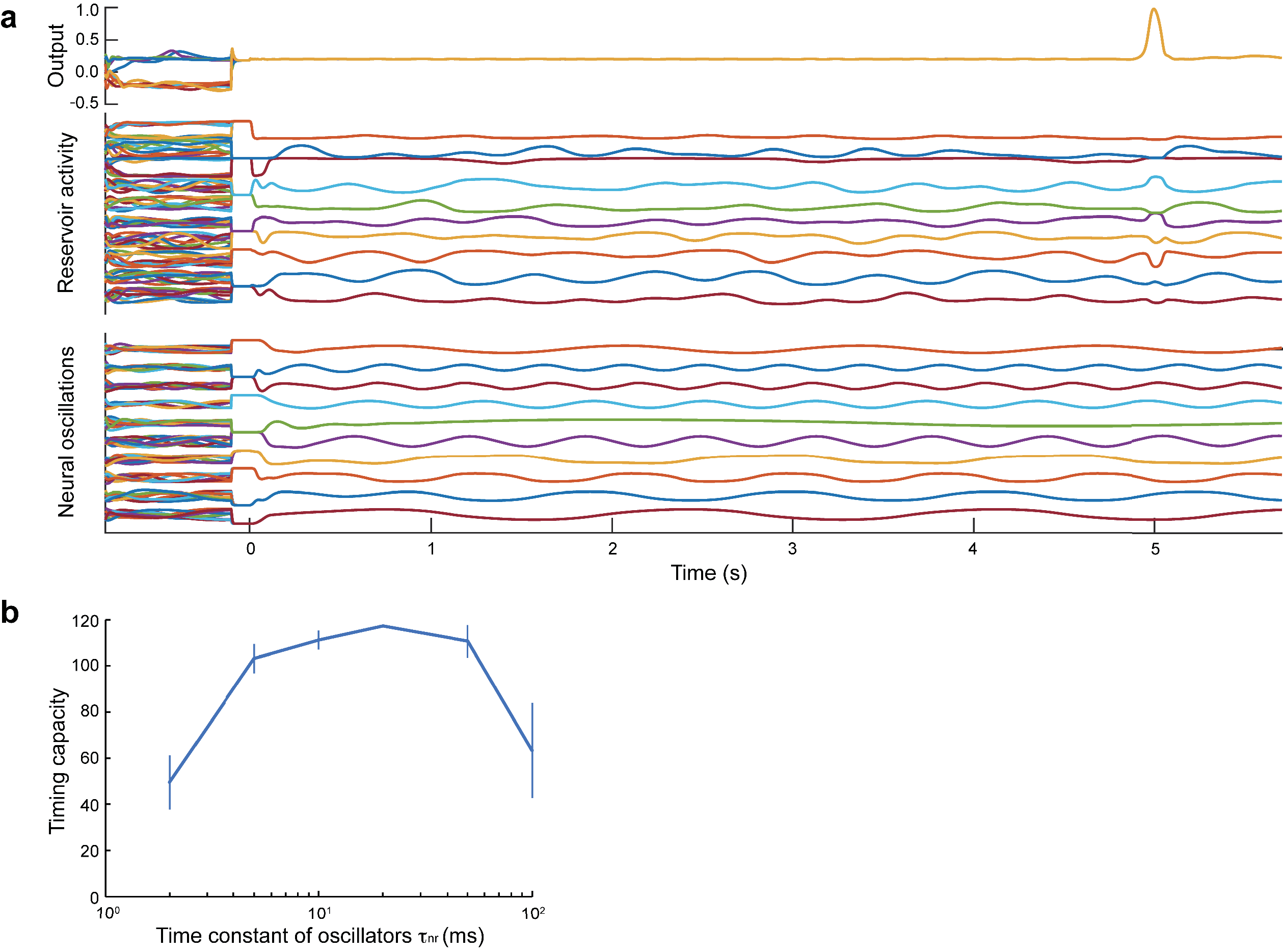}
		\caption{Results of neural oscillator-driven reservoir computing (ODRC) for the motor timing task.
			{\bf a} Trajectories of the output (top), reservoir activity of ten neural units (middle), and oscillations (bottom) in the 5-s task.
			These are the overlaid curves from ten test trials.
			{\bf b} Timing capacities when varying the time constant of oscillators.
			The performance was averaged over ten networks (mean $\pm$ s.d.).}
		\label{fig:neural_timing}
	\end{center}
\end{figure}

The neural ODRC obtained results similar to those of the sine ODRC.
The dynamics of the neural oscillators are shown at the bottom of Fig.~\ref{fig:neural_timing}a.
The time constant of the oscillators was set to 20 ms, which produced stable limit cycles at lower frequencies than the reservoir dynamics.
Their frequencies differed from each other.
The neural oscillations stabilized the reservoir activity, as shown in the middle of Fig.~\ref{fig:neural_timing}a.
As sine ODRC, the neural ODRC can generate a reliable output.
The frequencies of the oscillators depend on the time constant.
In this task, the performance peaked at 20 ms, and smaller (faster) and larger (slower) time constants resulted in performance degradation (Fig.~\ref{fig:neural_timing}b).
The peak performance was comparable to that of the sine ODRC with feedback in the $[0.1, 1]$ Hz frequency band (see Fig.~\ref{fig:sine_timing}b and d).

\subsubsection{Robustness against noise}
\begin{figure}[!tb]
	\begin{center}
		\includegraphics*[width=13.0cm]{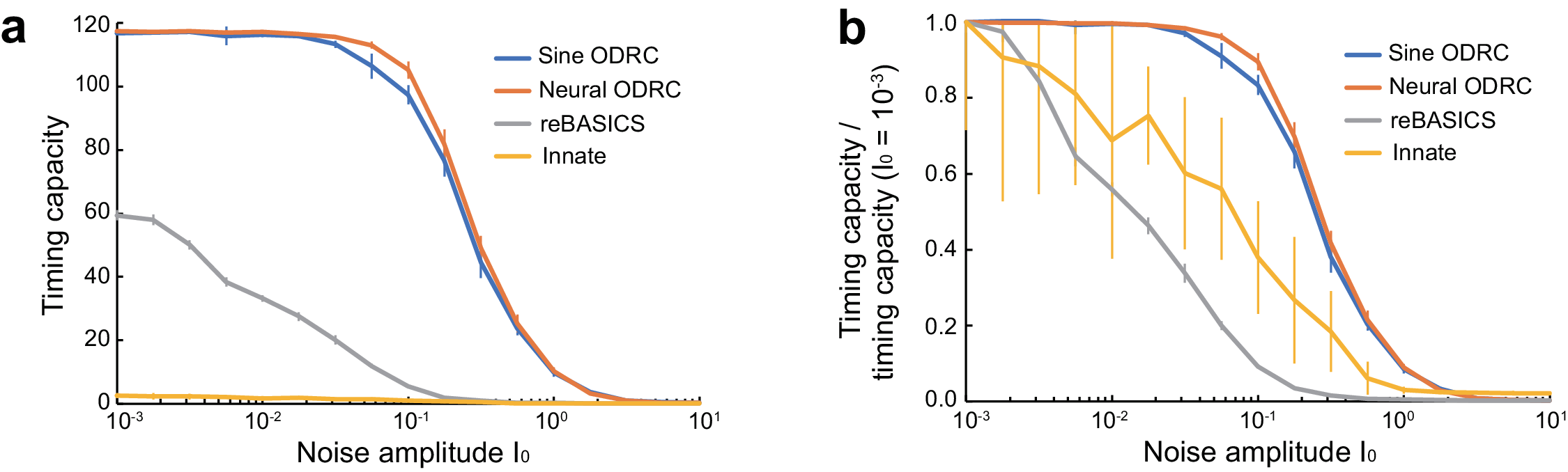}
		\caption{Robustness against noise in the motor timing task.
			The timing capacities were evaluated when Gaussian noise (amplitude $I_0$) was added to all reservoir neural units during training and testing.
			The performance of tabularhe sinusoidal oscillator-driven reservoir computing (sine ODRC), neural ODRC, reservoir of basal dynamics (reBASICS)~\cite{kawai2023learning}, and innate training~\cite{laje2013robust} was evaluated.
			The performance was averaged over ten networks (mean $\pm$ s.d.).
			{\bf a} Timing capacity under noise.
			{\bf b} Timing capacity normalized by the timing capacity when $I_0 = 10^{-3}$.}
		\label{fig:noise}
	\end{center}
\end{figure}

We examined the noise robustness of the ODRC with feedback during a motor timing task.
During training and testing, all neural units of the reservoirs received Gaussian noise with a mean of zero and standard deviation of $I_0$ (noise amplitude).
$I_0$ was set from $10^{-3}$ to $10$.
Fig.~\ref{fig:noise}a shows the timing capacities under noisy conditions.
The sine and neural ODRC, in which the oscillators did not contain noise, exhibited robustness against noise.
To compare the robustness of these with other methods, the timing capacities were normalized when $I_0 = 10^{-3}$ (Fig.~\ref{fig:noise}b).
The sine and neural ODRC exhibited better robustness than reBASICS~\citep{kawai2023learning} and an innate training method that explicitly trains for noise reduction~\citep{laje2013robust}.
This indicates that oscillation inputs can reduce the noise effects.

\subsection{Chaotic time-series prediction}
\subsubsection{Lorenz system}
The chaotic time-series prediction task targets a multi-dimensional chaotic time series.
First, we used a typical three-dimensional chaotic time series called the Lorenz system~\citep{lorenz1963deterministic}.
The task period was set from 1 s to 80 s.
$R^2$ was evaluated for each dimension and averaged over the dimensions.
We defined the Lorenz capacity as the area under the averaged $R^2$ curve for up to 80 s.

\begin{figure}[!tb]
	\begin{center}
		\includegraphics*[width=13.0cm]{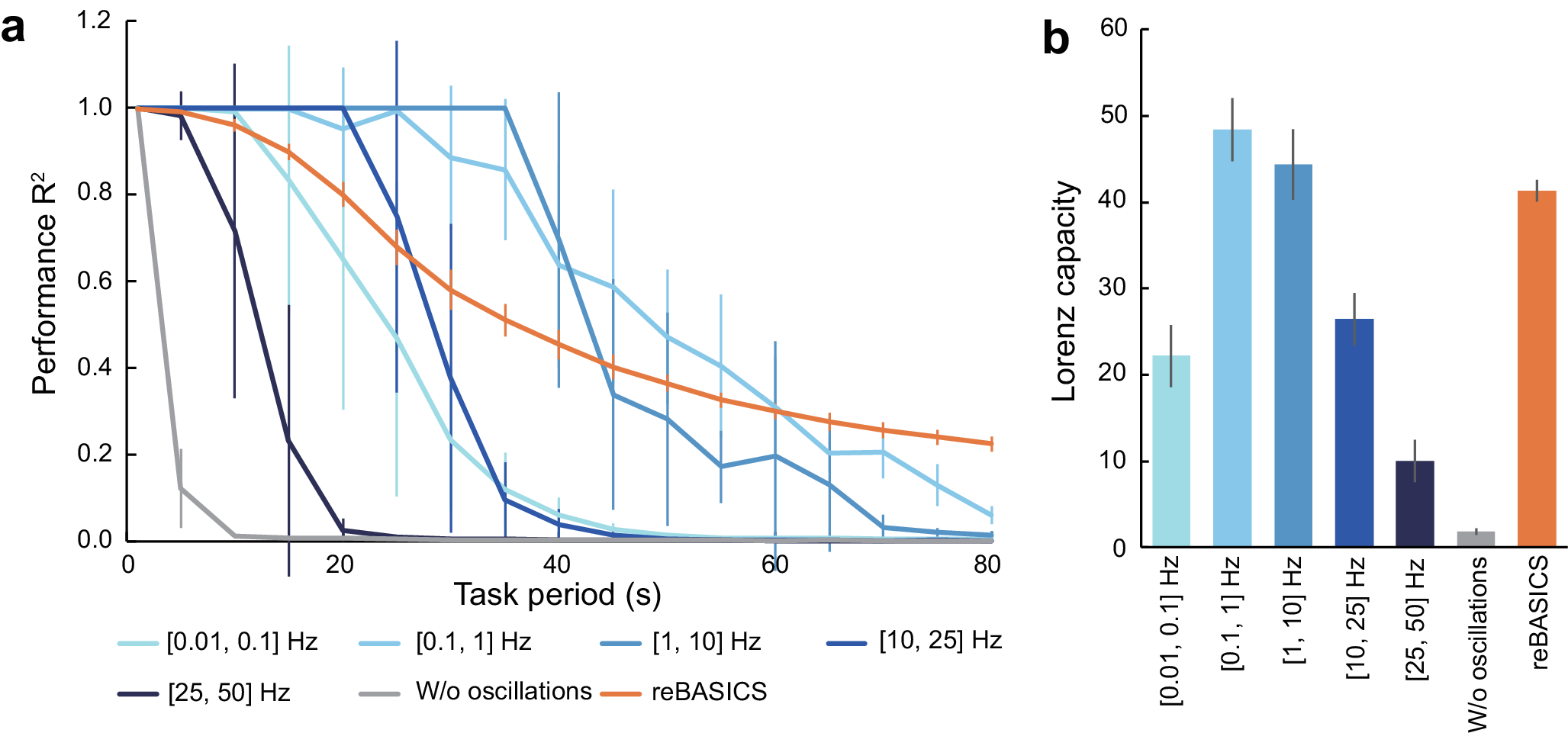}
		\caption{Results of sinusoidal oscillator-driven reservoir computing (ODRC) for the Lorenz time-series prediction task.
			The performance was averaged over ten networks (mean $\pm$ s.d.).
			{\bf a} Averaged performance $R^2$ for up to 80 s of the task periods, which is defined as the square of the correlation coefficient between the target time series and model outputs.
			The performance of the ODRC with various frequency bands, reservoir computing without (w/o) oscillations, and reservoir of basal dynamics (reBASICS)~\cite{kawai2023learning} was evaluated.
			{\bf b} Performance curves summarized as the areas under the curves, referred to as Lorenz capacities.}
		\label{fig:lorenz_capacity}
	\end{center}
\end{figure}

Fig.~\ref{fig:lorenz_capacity}a shows the averaged $R^2$ for the sine ODRC with various frequency bands, reservoir computing without oscillations, and the reBASICS~\citep{kawai2023learning}.
The performance is summarized in terms of Lorenz capacity in Fig.~\ref{fig:lorenz_capacity}b.
Similar to the results of the motor timing task, the sine ODRC at $[0.1, 1]$ Hz exhibited the best performance.
This suggests that the relatively low-frequency oscillator inputs were effective in reproducing the target time series, regardless of the task.

\begin{figure}[!tbhp]
	\begin{center}
		\includegraphics*[width=13.0cm]{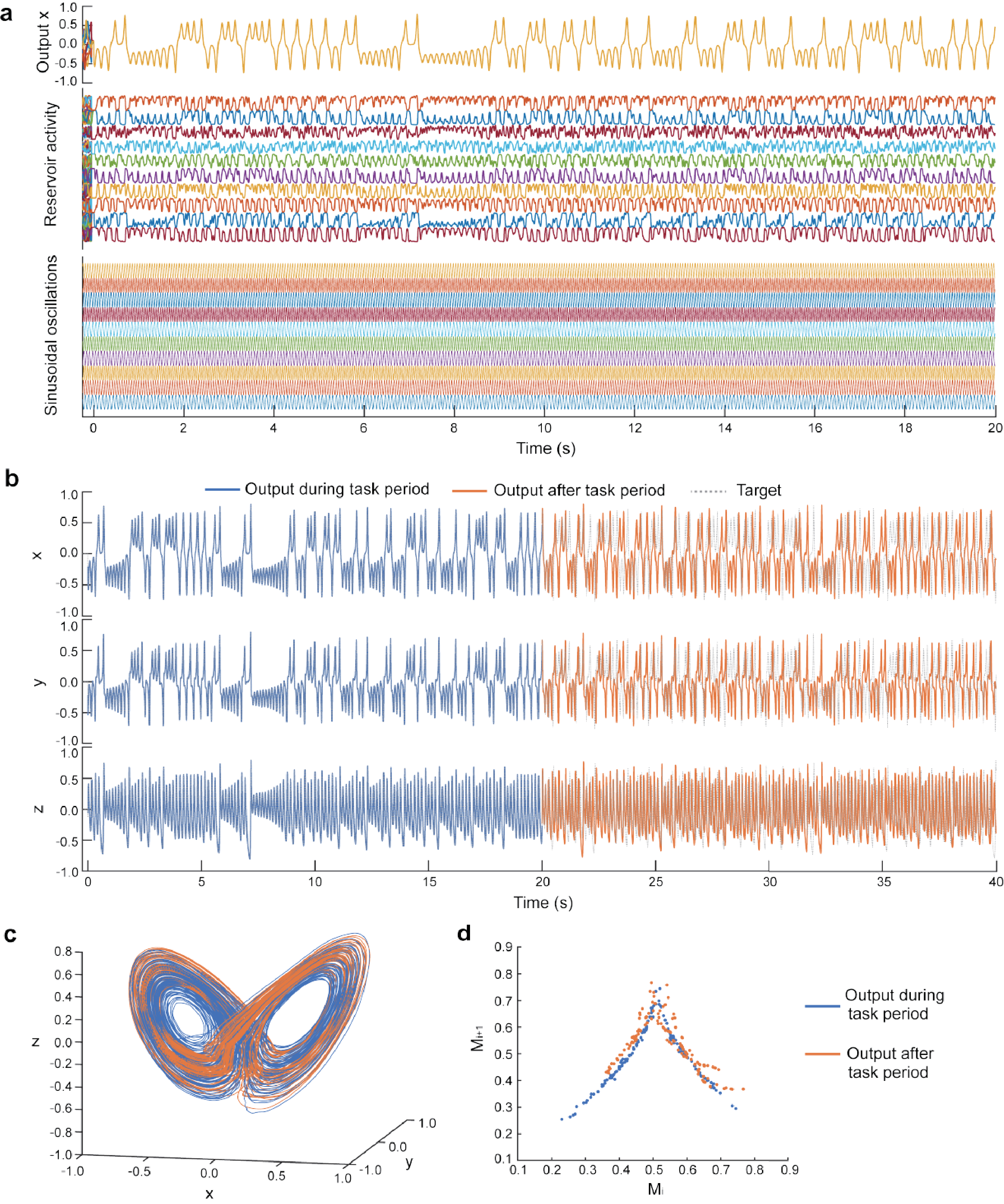}
		\caption{Example result of sinusoidal oscillation-driven reservoir computing (ODRC) for the Lorenz time-series prediction task.
			The ODRC had oscillator inputs in the $[10, 25]$ Hz frequency band.
			{\bf a} Trajectories of the output for $x$-coordinate (top), reservoir activity of ten neural units (middle), and oscillations (bottom) in the 20-s task.
			These are the overlaid curves from ten test trials.
			{\bf b} Three-dimensional output for 40 s after training for 20 s. The blue and orange curves indicate output during and after the task period of 20 s, respectively.
			The broken gray curve indicates the target Lorenz time series.
			{\bf c} Three-dimensional plot of the output trajectory.
			{\bf d} Return map of successive maxima of the output for $z$-coordinate.}
		\label{fig:lorenz}
	\end{center}
\end{figure}

As described later, low-frequency inputs, including $[0.1, 1]$ Hz, reduced the ability to generalize the target time series and henceforth results for $[10, 25]$ Hz are presented.
Fig.~\ref{fig:lorenz}a shows an example of the dynamics of the sine ODRC at $[10, 25]$ Hz for the 20-s task.
Oscillation inputs enabled reservoir activity to be complex and stable, that is, with low trial-to-trial variability.
Consequently, the ODRC could reproduce long-term complex time series, such as the Lorenz time series, without variability.

To examine the generalization capability of the ODRC, the test period was set to 40 s with a training period of 20 s.
In the test, the latter 20 s were the unexperienced periods.
If the ODRC can generate Lorenz-like trajectories during this period, it is considered to have generalized the target.
Fig.~\ref{fig:lorenz}b shows an example of the output trajectories for the sine ODRC at $[10, 25]$ Hz during the test period.
The ODRC generates a time series that matches the target with high accuracy during the task period, after which it generates a time series analogous to the Lorenz time series.
Reservoir computing without oscillations allowed the generalization of the target; however, it did not allow the reproduction of the target during the task period (\ref{appendix2}).

The three-dimensional plot of the output trajectory also demonstrates that the output represents the Lorenz attractor (Fig.~\ref{fig:lorenz}c).
In Fig.~\ref{fig:lorenz}d, we plotted the return map of the successive maxima of $z$.
We first located all local maxima of the output trajectory for $z$-coordinate in time order and denoted them $[M_1, M_2, \dots, M_n]$.
We then plotted consecutive pairs of those maxima $[M_i, M_{i+1}]$ ($i = 1, \dots n-1$) as dots.
The dots overlapped during and after the task period, forming a tent-like shape; therefore, the output trajectories exhibited characteristics of the Lorenz system.
Furthermore, the Lyapunov spectrum was used to characterize the output time series, revealing a Lyapunov spectrum similar to that of the original Lorenz system (\ref{appendix3}).

The generalization capability of the ODRC depends on the frequency band of the oscillators (\ref{appendix4}).
With frequency bands lower than $[10, 25]$ Hz, reproduction during the task period was possible; however, generalization after the task period was unsuccessful.
By contrast, with frequency bands higher than $[10, 25]$ Hz, generalization was possible; however, reproduction during the task period was unsuccessful.
Therefore, relatively high-frequency oscillations are effective for target generalization.

\subsubsection{R\"{o}ssler and Kuramoto-Sivashinsky (KS) systems}
\begin{figure}[!tbhp]
	\begin{center}
		\includegraphics*[width=13.0cm]{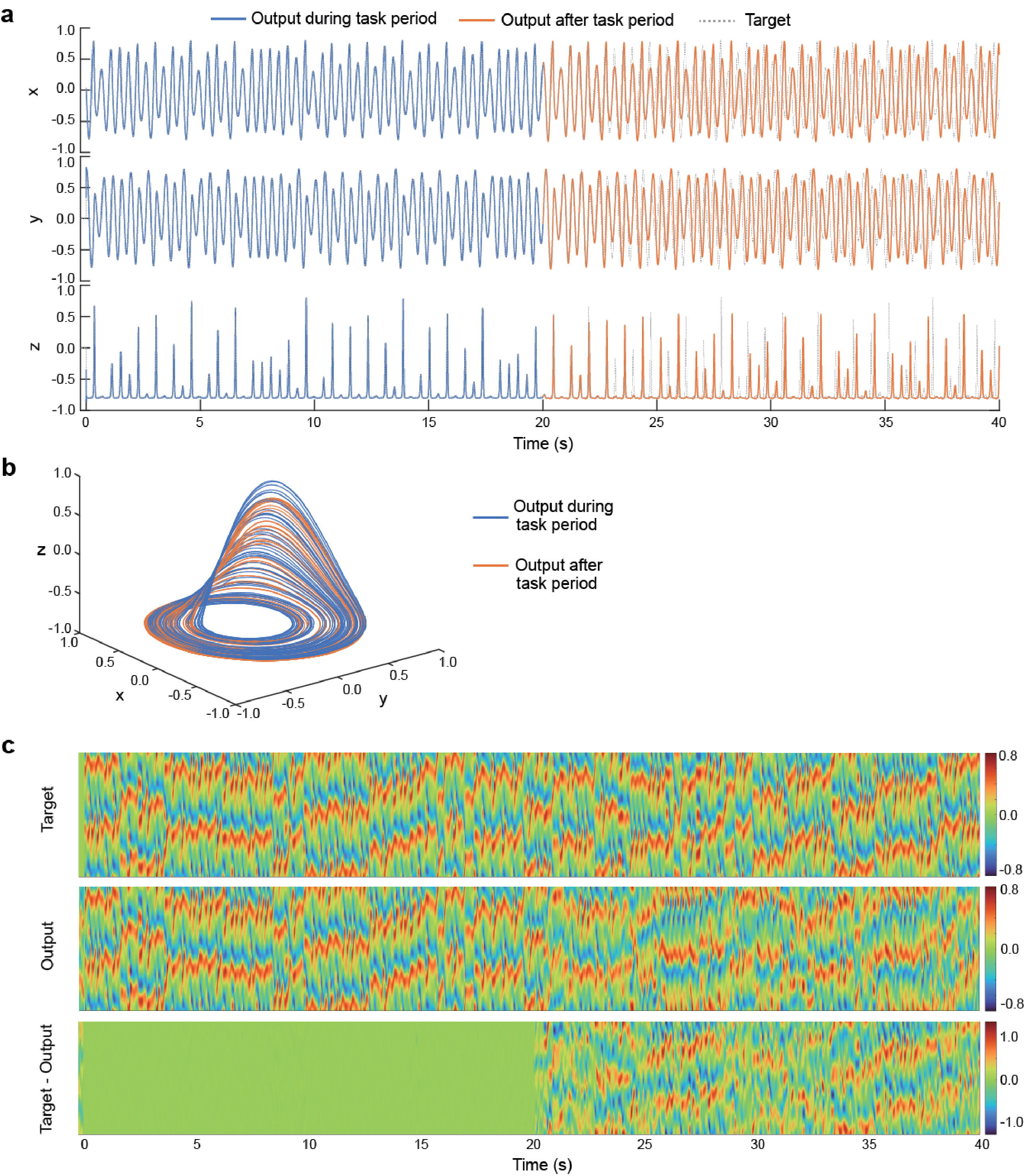}
		\caption{Example result of sinusoidal oscillation-driven reservoir computing (ODRC) for the R\"{o}ssler and Kuramoto-Sivashinsky (KS) time-series prediction tasks (20 s).
			{\bf a} Three-dimensional output for 40 s after training the R\"{o}ssler time series for 20 s.
			The blue and orange curves indicate the outputs during and after a task period of 20 s, respectively.
			The broken gray curve indicates the target R\"{o}ssler time series.
			{\bf b} Three-dimensional plot of the output trajectory for the R\"{o}ssler time-series prediction task.
			{\bf c} Contour plots of the target KS time series (top), ODRC output (middle), and prediction differences from the target for the output (bottom).}
		\label{fig:rossler+ks}
	\end{center}
\end{figure}

Subsequently, we evaluated the sine ODRC at $[10, 25]$ Hz using the R\"{o}ssler time series~\citep{rossler1976equation}, a three-dimensional chaotic time series.
As with the Lorenz time-series prediction task, the ODRC learned a 20-s R\"{o}ssler time series and generated a 40-s time series during testing (Fig.~\ref{fig:rossler+ks}a).
This example indicates that the target was reproduced with high accuracy during the task period, and a time series similar to the R\"{o}ssler time series was generated after the task period.
The three-dimensional plot also demonstrates that the output trajectory after the task period overlapped with that of the target (Fig.~\ref{fig:rossler+ks}b).

Finally, the same evaluation of the sine ODRC at $[10, 25]$ Hz was performed on a 64-dimentional KS system~\citep{kuramoto1978diffusion, sivashinsky1977nonlinear}, which is a spatiotemporally chaotic time series (Fig.~\ref{fig:rossler+ks}c).
During the task period, the error between the target and output was approximately zero, and after the task period, the output was similar to the target.

Similar evaluations were conducted with the neural ODRC in the Lorenz, R\"{o}ssler, and KS time-series prediction tasks, and results similar to the sine ODRC were obtained (\ref{appendix5}).
The time constant of the oscillator networks of the neural ODRC was 2 ms.
The neural ODRC reproduced the target time series for the 20-s task period and generated a time series analogous to the target thereafter.

\section{Discussion}
In this study, we proposed the ODRC, a reservoir computing model with feedback in which oscillations are fed into a reservoir network to stabilize the system and predict long-term time series.
Observations of the reservoir dynamics indicated that the inputs of the sinusoidal or neural oscillators suppressed the chaoticity or instability in the reservoir dynamics.
The reservoirs showed complex and stable activities, allowing them to represent the long-term time series.
The ODRC outperformed the conventional reservoir computing methods and better predicted the long-term motor timing and chaotic time-series prediction tasks.
Furthermore, the ODRC learned the generative rules behind the chaotic time series to produce a time series analogous to the target chaotic time series after the task period.
Therefore, the ODRC can accurately reproduce the target time series during the task period and generalize it after the task period.

An analysis of the parameter dependence of the ODRC revealed that oscillator inputs in relatively low-frequency bands, for example, $[0.1, 1]$ Hz, were effective in the motor timing task.
In the Lorenz time-series prediction task, performance in reproducing the target for the task period was maximal in the $[0.1, 1]$ Hz frequency band.
Therefore, relatively low-frequency inputs are useful for the accurate reproduction and memorization of the target time series.

\begin{figure}[tb]
	\begin{center}
		\includegraphics*[width=10.0cm]{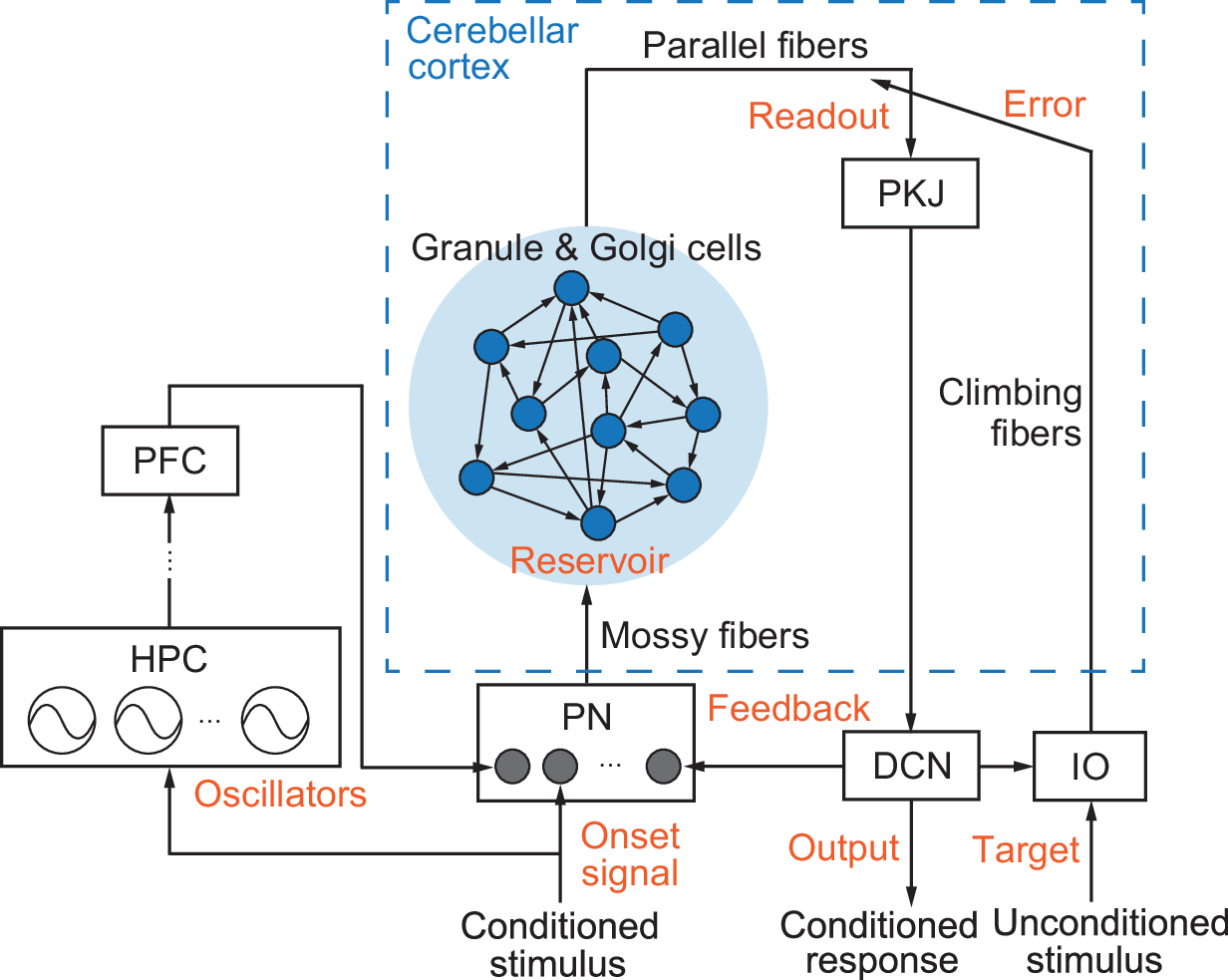}
		\caption{Simplified circuit model for trace eyeblink conditioning.
      The conditioned stimulus (an onset signal) is fed into the granule-layer network through the pontine nuclei (PN).
      Purkinje cells (PKJ) receive the network outputs and project their outputs to the deep cerebellar nuclei (DCN), which are closely related with the motor output, i.e., the conditioned response.
      The outputs are also fed into the inferior olive (IO) to create an error signal that modulates the synapses of the Purkineje cells through long-term depression.
      The Purkinje-cell outputs are also projected to the PN, forming a feedback loop~\citep{xiao2023positive}.
      Oscillations from the hippocampus (HPC) are fed into the PN through the prefrontal cortex (PFC)~\citep{hoffmann2015harnessing}, which sustain activity of the granule-layer network during the trace interval.}
		\label{fig:hippobellum}
	\end{center}
\end{figure}

In biological timing learning, particularly in trace eyeblink conditioning, theta oscillations (3--7 Hz) in the hippocampus are required~\citep{solomon1986hippocampus, nokia2009hippocampal, hoffmann2009cerebellar, berry2011hippocampal}.
If reservoir computing is regarded as a model of the cerebellar cortex~\citep{yamazaki2007cerebellum, rossert2015edge, tokuda2021chaos}, granule cells receive cortical inputs through mossy fibers, and granule cells and Golgi cells form a recurrent neural network, i.e., reservoir network (Fig.~\ref{fig:hippobellum}).
Outputs of the granule cells are sent to Purkinje cells to produce the system output.
The plasticity of the Purkinje cells is regarded as readout training.
Recently, a feedback loop from the Purkinje cell outputs (efference copy) to the granule layer through the deep cerebellar nuclei and the pontine nucleus has been found to facilitate classical delayed eyeblink conditioning in rats~\citep{xiao2023positive}.
A possible model is that low-frequency inputs from the hippocampus drive the granule cell networks to sustain the stimulus-induced activity during the trace interval.
The hippocampus activates the networks even if there is a stimulus-free interval between stimuli, thus achieving motor timing learning.
Our results suggest that low-frequency inputs, including theta oscillations, are effective for the cerebellar learning because they sustain and stabilize the cerebellar activity.
In addition, low-frequency oscillations affect activity of neurons with a time constant more than high-frequency oscillations.
However, the current ODRC's neuron model is an abstract firing-rate model and requires further investigation using more biologically plausible spiking neural networks to explore effective frequencies for timing learning.

The ODRC has implications for other timing-learning models.
A timing-learning model in cortico-striatal circuits predicts timing by recording and referring to the firing states of cortical oscillators at that timing~\citep{miall1989storage, matell2004cortico}.
The ODRC extends this model to be more computationally powerful.
The reservoir is regarded as a neural network of the striatum~\citep{dominey1995complex, hinaut2013real, kawai2023spatiotemporal} that receives oscillatory signals from the cortex.
These cortical oscillations stabilize striatal reservoir activity and induce the complex activity.
Dopamine modifications in the striatum facilitate readout learning, allowing the learning of long-term timing and temporal patterns.

The idea of inputting sinusoidal signals to a learning machine has also been used in convolutional neural networks~\citep{gehring2017convolutional} and Transformers~\citep{vaswani2017attention}.
When series data are applied to these models, the data are processed simultaneously rather than sequentially.
Therefore, these methods require a mechanism to represent the order or sequence of the input data or tokens.
To provide temporal information of the input data as ``time stamps,'' low-frequency sinusoidal signals accompany the data, which is called positional encoding.
Recently, it was reported that introducing positional encoding into various recurrent neural networks improves their performance~\citep{morita2024positional}.
Recurrent neural networks should represent temporal information internally, and positional encoding may assist in temporal representation.
It is assumed that a reason for the high performance of the ODRC is the timestamping of the input time series by positional encoding.

Low frequencies are effective for target reproduction, whereas relatively high frequencies are effective for target generalization.
As the time constant of the reservoir neuron model was 10 ms, high-frequency inputs stabilized their activity and did not impair the reservoir's inherent dynamics without significantly affecting the increase or decrease in neural activity.
It is therefore possible that the ODRC trains outputs from stabilized reservoir activity rather than overfitted outputs for specific oscillatory inputs, which might improve the generalization ability.

The framework of the ODRC is simple and has wide applicability.
The ODRC can be applied to physical reservoir computing~\citep{tanaka2019recent}, which is promising as energy-efficient computing systems, as well as neural reservoir computing.
Existing physical reservoir computing often uses nonlinear oscillators as reservoir network units~\citep{dion2018reservoir, tsunegi2019physical, govia2021quantum}.
Using nonlinear oscillators for network stabilization and complexity is expected to improve performance.
The ODRC will contribute to the understanding of the role of oscillations in the brain and the development of energy-saving neuromorphic computers.

\section{Method}
\subsection{Sine ODRC}
The reservoir comprises $N$ neural units.
At time $t$, the state vector $\mathbf{x}(t) = (x_1, x_2, \dots, x_N)^\top$ of the neural units is given as
\begin{eqnarray}
	\label{eq:1}
	\tau \frac{\mathrm{d}\mathbf{x}(t)}{\mathrm{d}t} &=& -\mathbf{x}(t) + \mathbf{W} \mathbf{r}(t) + \mathbf{W}_\mathrm{os} \mathbf{o}(t) + \mathbf{W}_\mathrm{in} s(t) + \mathbf{W}_\mathrm{fb} \mathbf{y}_\mathrm{fb}(t) + \mathbf{I}_\mathrm{noise},\\
	\label{eq:2}
	\mathbf{r} (t) &=& \tanh(\mathbf{x}(t)),
\end{eqnarray}
where $\tau$ denotes the time constant, $\mathbf{I}_\mathrm{noise}$ denotes the noise term, and $\mathbf{W}$, $\mathbf{W}_\mathrm{os}$, $\mathbf{W}_\mathrm{in}$, and $\mathbf{W}_\mathrm{fb}$ denote the reservoir recurrent weight matrix, oscillator input weight matrix, onset signal input weight vector, and feedback weight matrix, respectively.
Variables $\mathbf{o}(t)$, $s(t)$, and $\mathbf{y}_\mathrm{fb}(t)$ denote the oscillation inputs, an onset signal, and the output feedback signals, respectively, at time $t$.
The initial values of $\mathbf{x}(t)$ are determined randomly with a uniform distribution in the range $[1, -1]$.

$\mathbf{W}$ is an $N \times N$ weight matrix, and each component has a non-zero value with a connection probability $p$.
Its non-zero values are drawn from a Gaussian distribution with a mean of zero and standard deviation of $g/\sqrt{pN}$, where $g$ is a gain of the reservoir weights.

The oscillators are $N_\mathrm{os}$ sinusoids, with vector $\mathbf{o} (t) = (\sin(2 \pi f_1 t + \phi_1), \sin(2 \pi f_2 t + \phi_2), \dots, \sin(2 \pi f_{N_\mathrm{os}} t + \phi_{N_\mathrm{os}}))^\top$, where $f_i$ and $\phi_i$ denote the frequency drawn from a uniform distribution in the interval $[f_\mathrm{min}, f_\mathrm{max}]$ and the random initial phase.
These parameters are fixed across trials.
The oscillations are fed into the reservoir units through an $N_\mathrm{os} \times N$ weight matrix $\mathbf{W}_\mathrm{os}$.
Its elements are drawn from a Gaussian distribution with a mean of zero and standard deviation of $g_\mathrm{os}/\sqrt{N_\mathrm{os}}$, where $g_\mathrm{os}$ denotes a gain of the oscillation input weights.

The onset signal $s(t)$ is a scalar single pulse used to suppress the initial value dependence of the reservoir network.
From $-50$ ms to $0$ ms, $s(t)$ is 1, and at other times it is 0.
This pulse is fed into the reservoir units through the weight vector $\mathbf{W}_\mathrm{in}$ of size $N$.
The elements are drawn from a Gaussian distribution with a mean of zero and standard deviation of $g_\mathrm{in}$, where $g_\mathrm{in}$ denotes the gain of the onset signal input weights.

The reservoir states $\mathbf{x}(t)$ are integrated into the readout using the following equation to produce $N_\mathrm{ro}$-dimensional output $\mathbf{y}(t) = (y_1, y_2, \dots, y_{N_\mathrm{ro}})$:
\begin{equation}
	\label{eq:3}
	\mathbf{y}(t) = \mathbf{W}_\mathrm{ro}(t) \mathbf{r}(t),
\end{equation}
where $\mathbf{W}_\mathrm{ro}(t)$ denotes an $N \times N_\mathrm{ro}$ readout weight matrix modulated by recursive least squares.
The initial value of $\mathbf{W}_\mathrm{ro}(t)$ is the zero matrix.
The readout output $\mathbf{y}(t)$ is fed back into the reservoir units as $\mathbf{y}_\mathrm{fb}(t)$ through the $N_\mathrm{ro} \times N$ feedback weight matrix $\mathbf{W}_\mathrm{fb}$.
The elements are drawn from a Gaussian distribution with a mean of zero and standard deviation of $g_\mathrm{fb} / \sqrt{N_{\mathrm{fb}}}$, where $g_\mathrm{fb}$ denotes a gain of the feedback weights.

In the simulations, the numerical solutions of equation~(\ref{eq:1}) are obtained using the Euler method:
\begin{equation}
	\label{eq:4}
	\mathbf{x}(t + \Delta t) = \left( 1 - \frac{\Delta t}{\tau} \right) \mathbf{x}(t) + \frac{\Delta t}{\tau} \left(\mathbf{W} \mathbf{r}(t) + \mathbf{W}_\mathrm{os} \mathbf{o}(t) + \mathbf{W}_\mathrm{in} s(t) + \mathbf{W}_\mathrm{fb} \mathbf{y}_\mathrm{fb}(t) + \mathbf{I}_\mathrm{noise} \right).
\end{equation}
The simulation step size $\Delta t$ is set to 1 ms.

The weight matrix $\mathbf{W}_\mathrm{ro} (t)$ is trained to minimize the error $\mathbf{e}(t)$ between $\mathbf{y}(t)$ and target $\mathbf{d} (t)$ using the recursive least squares method, which is an online learning method.
At time $t$, $\mathbf{W}_\mathrm{ro} (t)$ is updated as follows:
\begin{eqnarray}
	\label{eq:5}
	\mathbf{W}_\mathrm{ro} (t + \Delta t) &=& \mathbf{W}_\mathrm{ro} (t) - \mathbf{e}(t) \mathbf{P}(t) \mathbf{r}(t),\\
	\label{eq:6}
	\mathbf{e}(t) &=& \mathbf{y}(t) - \mathbf{d}(t),
\end{eqnarray}
where $\mathbf{P}(t)$ is an $N \times N$ matrix, which is updated as
\begin{equation}
	\label{eq:7}
	\mathbf{P}(t + \Delta t) = \mathbf{P}(t) - \frac{\mathbf{P}(t) \mathbf{r}(t) \mathbf{r}^\top(t) \mathbf{P}(t)}{1 + \mathbf{r}^\top(t) \mathbf{P}(t) \mathbf{r}(t)}.
\end{equation}
The initial value of $\mathbf{P}(t)$ is given as $(1/\alpha)\mathbf{I}$, where $\mathbf{I}$ denotes an identity matrix, and $\alpha$ is a constant.
The recursive least squares (equations~(\ref{eq:5})--(\ref{eq:7})) are applied once every two steps, and $\Delta t$ is set to 2 ms.

The simulation starts at $-250$ ms, and the training period lasts from 1 ms until the end of the task.
The training is repeated ten times, and performance is evaluated during the untrained test period.

\subsection{Neural ODRC}
The neural ODRC is identical to the sine ODRC, except for the oscillator signal $\mathbf{o}(t)$.
This signal is generated from $N_\mathrm{os}$ random neural networks of size $N_\mathrm{nr}$.
The neuron model of the oscillator networks is similar to that of the reservoir network.
The state $\mathbf{x}_i (t) = (x_{i1}, x_{i2}, \dots, x_{iN_\mathrm{nr}})^\top$ of the $i$th oscillator network at time $t$ is given as
\begin{eqnarray}
	\label{eq:8}
	\tau_\mathrm{nr} \frac{\mathrm{d}\mathbf{x}_i(t)}{\mathrm{d}t} &=& -\mathbf{x}_i(t) + \mathbf{W}^\mathrm{nr} \mathbf{r}_i(t) + \mathbf{W}^\mathrm{nr}_\mathrm{in} s(t),\\
	\label{eq:9}
	\mathbf{r}_i(t) &=& \tanh(\mathbf{x}_i(t)),
\end{eqnarray}
where $\tau_\mathrm{nr}$ denotes a time constant, and $\mathbf{W}^\mathrm{nr}$ and $\mathbf{W}^\mathrm{nr}_\mathrm{in}$ denote an $N_\mathrm{nr} \times N_\mathrm{nr}$ recurrent weight matrix and onset signal input weight vector of size $N_\mathrm{nr}$, respectively.
The elements of $\mathbf{W}^\mathrm{nr}$ and $\mathbf{W}^\mathrm{nr}_\mathrm{in}$ are drawn from Gaussian distributions with a mean of zero and standard deviation of $g_\mathrm{nr}/\sqrt{pN_\mathrm{nr}}$ and $g_\mathrm{in}$, respectively, where $g_\mathrm{nr}$ denotes a gain of the recurrent weights of the oscillator networks.
The oscillation signal is obtained as the state of a unit randomly chosen from each network.
If the oscillatory signal converges to a fixed point, $\mathbf{W}^\mathrm{nr}$ is resampled.

If $N_\mathrm{nr}$ is large and $g_\mathrm{nr} > 1$, the network activity becomes self-sustained but chaotic.
Setting $N_\mathrm{nr}$ small makes the network activity stable and oscillatory; that is, the networks generate limit cycles and tori, which are intermediate states between convergence to a zero fixed point and chaotic behavior~\cite{sompolinsky1988chaos, doyon1994bifurcations, kawai2023learning}.

\subsection{Chaotic time series}
We used the Lorenz, R\"{o}ssler, and KS systems as the target signals.
The Lorenz system is given by
\begin{eqnarray}
	\frac{\mathrm{d}x(t)}{\mathrm{d}t} &=& -\sigma (x(t) - y(t)),\\
	\frac{\mathrm{d}y(t)}{\mathrm{d}t} &=& x(t) (\rho - z(t)) - y(t),\\
	\frac{\mathrm{d}z(t)}{\mathrm{d}t} &=& x(t) y(t) - \beta z(t),
\end{eqnarray}
where $\sigma=10$, $\rho=28$, and $\beta=8/3$.
The R\"{o}ssler system is given by
\begin{eqnarray}
	\frac{\mathrm{d}x(t)}{\mathrm{d}t} &=& - y(t) - z(t),\\
	\frac{\mathrm{d}y(t)}{\mathrm{d}t} &=& x(t) + ay(t),\\
	\frac{\mathrm{d}z(t)}{\mathrm{d}t} &=& b + x(t) z(t) - c z(t),
\end{eqnarray}
where $a=0.2$, $b=0.2$, and $c=5.7$.
The numerical solutions of the Lorenz and R\"{o}ssler equations were obtained using the fourth-order Runge-Kutta method with a step size of 0.001, where $x(0)=0.1$, $y(0)=0$, and $z(0)=0$.
They were then down-sampled to 1/5 and 1/15 of their lengths.

KS system for $u(x, t)$ is given by
\begin{equation}
	\label{eq:16}
	\frac{\partial u}{\partial t} = - u \frac{\partial u}{\partial x} - \frac{\partial^2u}{\partial^2x} - \frac{\partial^4u}{\partial^4x}.
\end{equation}
Periodic boundary conditions, $u(x, t) = u(x+L, t)$ were applied, where $L$ ($= 22$) is the domain size ($0 \leq x \leq L$).
Equation~(\ref{eq:16}) was numerically integrated on a grid of 64 equally spaced points with a step size of 0.1 to obtain 64-dimensional time-series data.

In each system, the solution time series was normalized to $[-0.8, 0.8]$ in magnitude.
The first 3000 steps were discarded during the burn-in period.
One time step was regarded as 1 ms to obtain the target signal $\mathbf{d} (t)$.

\subsection{Parameter settings}
The parameter values in Table~\ref{table} were used in all the simulations.
For the motor-timing task, we set that $N_\mathrm{ro} = 1$, and $N = 400$.
In the chaotic time-series prediction task, we used reservoirs with $N = 3000$.
$N_\mathrm{ro} = 3$ for the Lorenz and R\"{o}ssler systems, and $N_\mathrm{ro} = 64$ for the KS system.

\begin{table}[t]
	\centering
	\caption{Parameter settings}
	\begin{tabular}{llr}
		Parameter & Description & Value \\ \hline
		$N_\mathrm{os}$ & Number of oscillators & 10 \\
		$N_\mathrm{nr}$ & Number of neural units of oscillator networks & 100 \\
		$g$ & Gain of reservoir weights & 1.5 \\
		$g_\mathrm{os}$ & Gain of oscillation input weights & 0.5 \\
		$g_\mathrm{in}$ & Gain of onset signal input weights & 5 \\
		$g_\mathrm{fb}$ & Gain of feedback weights & 3 \\
		$g_\mathrm{nr}$ & Gain of oscillator network weights & 1.2 \\
		$p$ & Connection probability & 0.1 \\
		$\tau$ & Time constant & 10 ms\\
		$\alpha$ & Initial value for recursive least squares & 1\\
		\hline
	\end{tabular}
	\label{table}
\end{table}

For comparison, we also evaluated the performance of reBASICS~\citep{kawai2023learning}, which comprises $M$ random network modules, with each module containing $N_m$ neural units.
We set $N_m = 100$, and $M = 400$ for all tasks.
Thus, the total number of neural units was 40,000.
The other parameters follow ref.~\citep{kawai2023learning}.
In the motor-timing task, the performance of the innate training method~\citep{laje2013robust} with 400 neural units was also evaluated.
The other parameters follow ref.~\citep{laje2013robust}.

\appendix
\newpage
\section{Appendix}
\subsection{Recurrent weight gain}
\label{appendix1}
\begin{figure}[!tb]
	\begin{center}
		\includegraphics*[width=8.0cm]{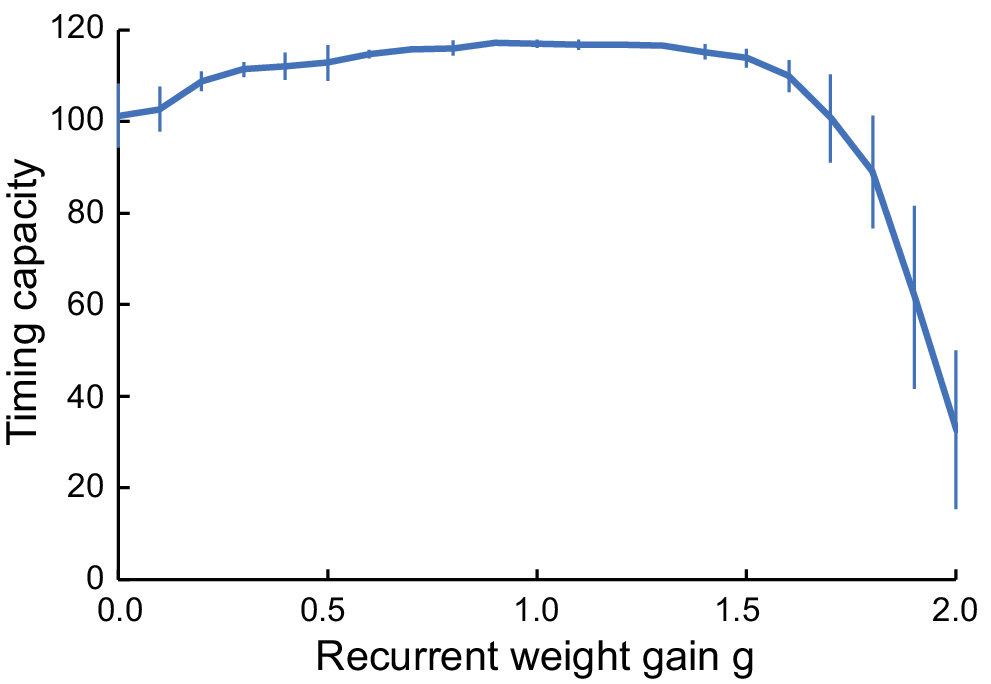}
		\caption{Timing capacities of the sinusoidal oscillator-driven reservoir computing when varying the gain of reservoir weights for the motor timing task.}
		\label{fig:sup_g}
	\end{center}
\end{figure}

We examined the sensitivity of the sine ODRC to the gain of reservoir weights in the motor timing task.
Fig.~\ref{fig:sup_g} shows the timing capacities versus the gain, indicating that $1.0 \leq g \leq 1.5$ yielded the optimum performance.

\subsection{Reservoir computing without oscillations}
\label{appendix2}
\begin{figure}[!tbhp]
	\begin{center}
		\includegraphics*[width=13.0cm]{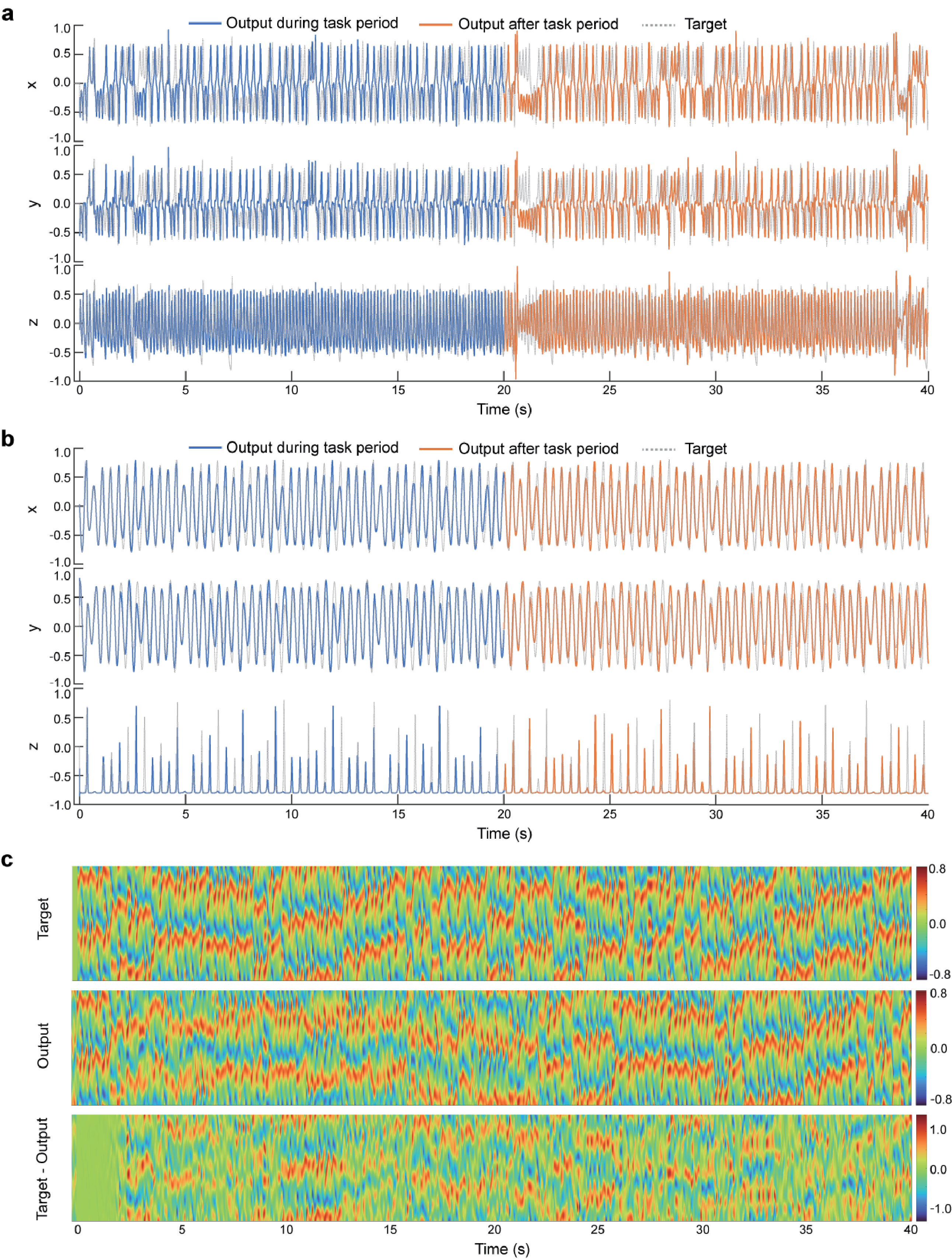}
		\caption{Example results of reservoir computing without oscillations for the chaotic time-series prediction tasks.
			{\bf a} Lorenz time-series prediction.
			{\bf b} R\"{o}ssler time-series prediction.
			{\bf a}, {\bf b} The blue, orange, and gray broken curves indicate outputs during and after the task period of 20 s and targets, respectively.
			{\bf c} Contour plots of the target Kuramoto--Sivashinsky time series (top), the output of the reservoir computing (middle), and the prediction differences from the target for the output (bottom).}
		\label{fig:sup_without}
	\end{center}
\end{figure}

Fig.~\ref{fig:sup_without} shows the outputs of the reservoir computing without oscillations in the chaotic time-series prediction.
The task duration was 20 s.
In each chaotic system, although the outputs failed to reproduce the target during the task period, they were analogous to the target throughout all simulation periods.

\subsection{Lyapunov spectrum analysis}
\label{appendix3}
One feature of chaotic dynamical systems is their sensitive dependence on initial conditions, where small differences in initial values exponentially increase over time and exhibit very different behaviors.
The exponent of expansion or contraction is referred to as the Lyapunov exponent.
The positive and negative Lyapunov exponents indicate orbital instability and stability, respectively.
In an $N$-dimensional dynamical system, the Lyapunov spectrum is the Lyapunov exponents of $N$ in increasing order.
If the Lyapunov spectra of the original target trajectory and reservoir outputs are similar, then the properties of the outputs are similar to those of the original dynamical system.
We use the Sano--Sawada method~\citep{sano1985measurement} to estimate the Lyapunov spectrum from the observed time series.
We estimated only the three-dimensional Lyapunov spectrum for the KS system using the three dimensions of the KS system because the dimensions of the KS time series were significantly large to accurately estimate the Lyapunov spectrum.

\begin{figure}[!tbhp]
	\begin{center}
		\includegraphics*[width=13.0cm]{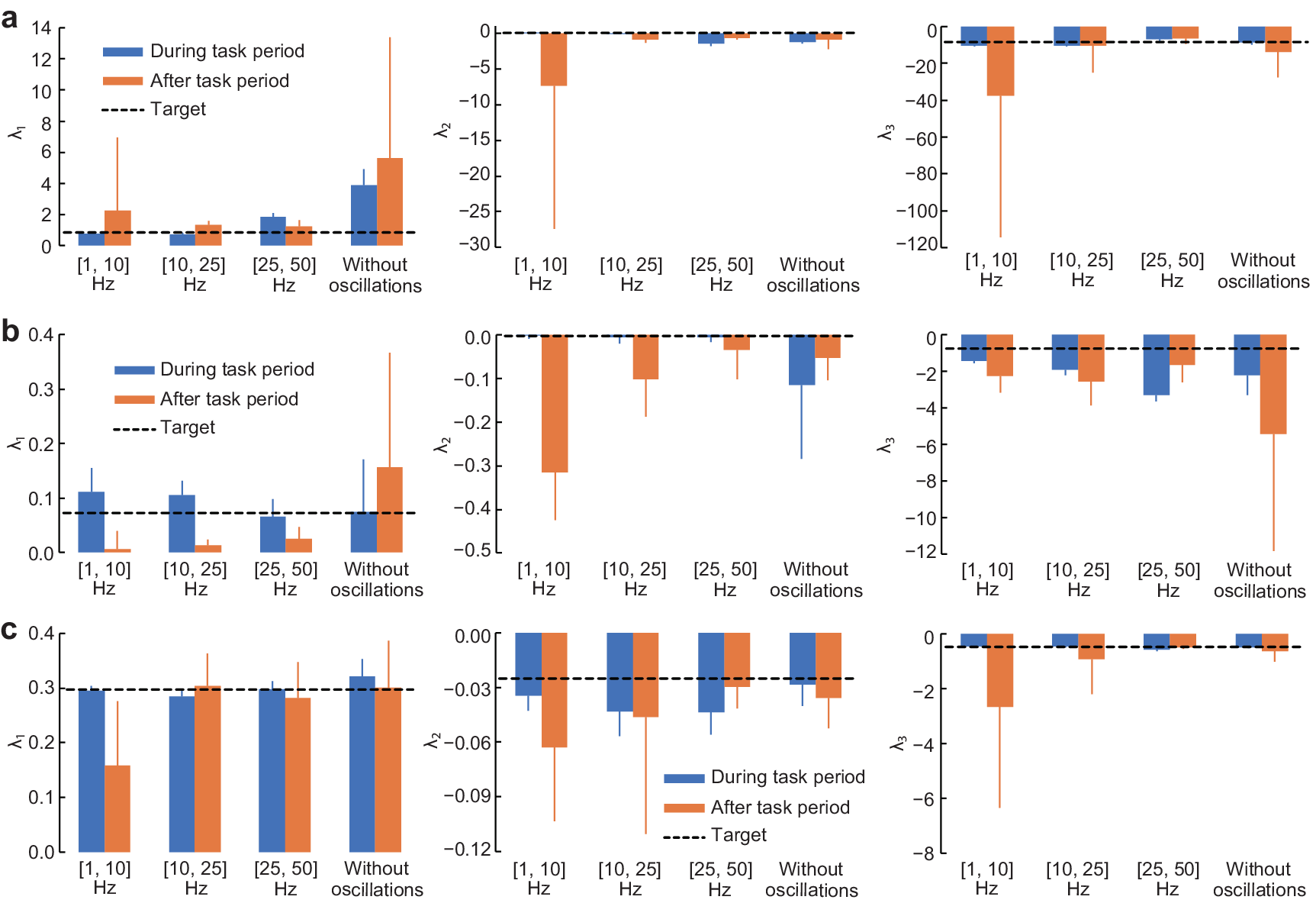}
		\caption{Estimated Lyapunov spectrum for sinusoidal oscillation-driven reservoir computing (ODRC).
			The horizontal broken lines indicate the Lyapunov exponents of the target.
			The blue and orange bars indicate the Lyapunov exponents of the outputs of the sinusoidal ODRC during and after the task period (20 s), respectively.
			The Lyapunov exponents were averaged over ten networks (mean $\pm$ s.d.).
			{\bf a}, {\bf b}, {\bf c} Lyapunov spectra for the Lorenz, R\"{o}ssler, and Kuramoto--Sivashinsky time-series predictions, respectively.}
		\label{fig:sup_lyapunov_spectrum}
	\end{center}
\end{figure}

Fig.~\ref{fig:sup_lyapunov_spectrum} shows the Lyapunov spectrum of the sine ODRC outputs for the Lorenz, R\"{o}ssler, and KS time-series predictions.
If the Lyapunov exponents were close to those of the target (broken lines), the outputs exhibited good reproduction and generalization.
The Lyapunov exponents of the ODRC outputs during the 20-s task period (blue bars) were close to the target in all frequency bands, suggesting that the outputs could replicate the target.
However, the Lyapunov exponents after the period (orange bars) depended on the frequency band: its outputs using $[10, 25]$ Hz and $[25, 50]$ Hz were closer to those of the targets than those using $[1, 10]$ Hz.
This result suggests that ODRC using high frequencies had the better generalizability to the target than those using low frequencies.

\subsection{Sine ODRC with different frequencies}
\label{appendix4}
\begin{figure}[!tbhp]
	\begin{center}
		\includegraphics*[width=13.0cm]{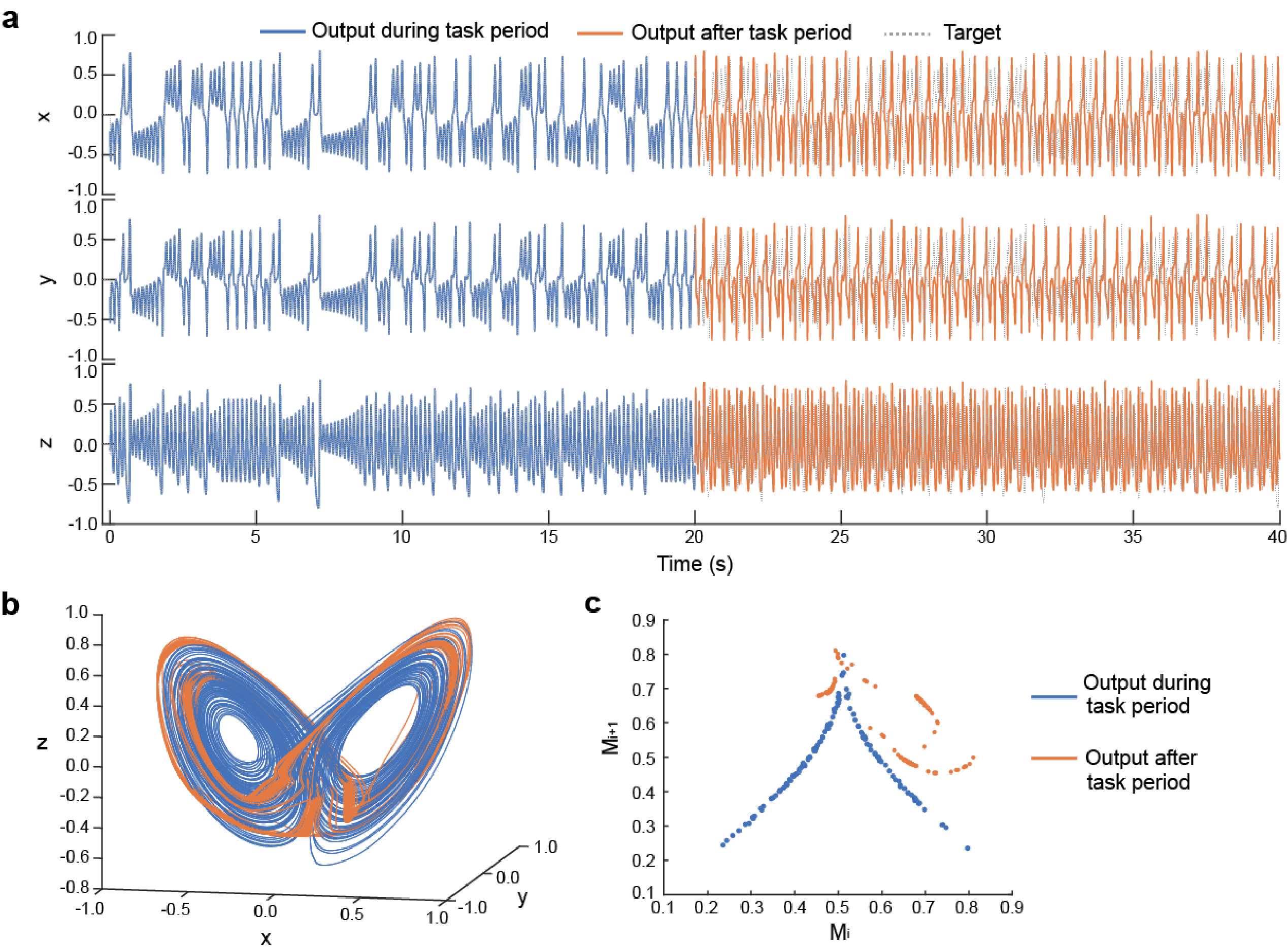}
		\caption{Example result of sinusoidal oscillation-driven reservoir computing with $[1, 10]$ Hz for the Lorenz time-series prediction task.
			{\bf a} Three-dimensional output for 40 s after training for 20 s. The blue and orange curves indicate output during and after the task period of 20 s, respectively.
			The broken gray curve indicates the target Lorenz time series.
			{\bf b} Three-dimensional plot of the output trajectory.
			{\bf c} Return map of successive maxima of the output for $z$-coordinate.}
		\label{fig:sup_5-10}
	\end{center}
\end{figure}

\begin{figure}[!tbhp]
	\begin{center}
		\includegraphics*[width=13.0cm]{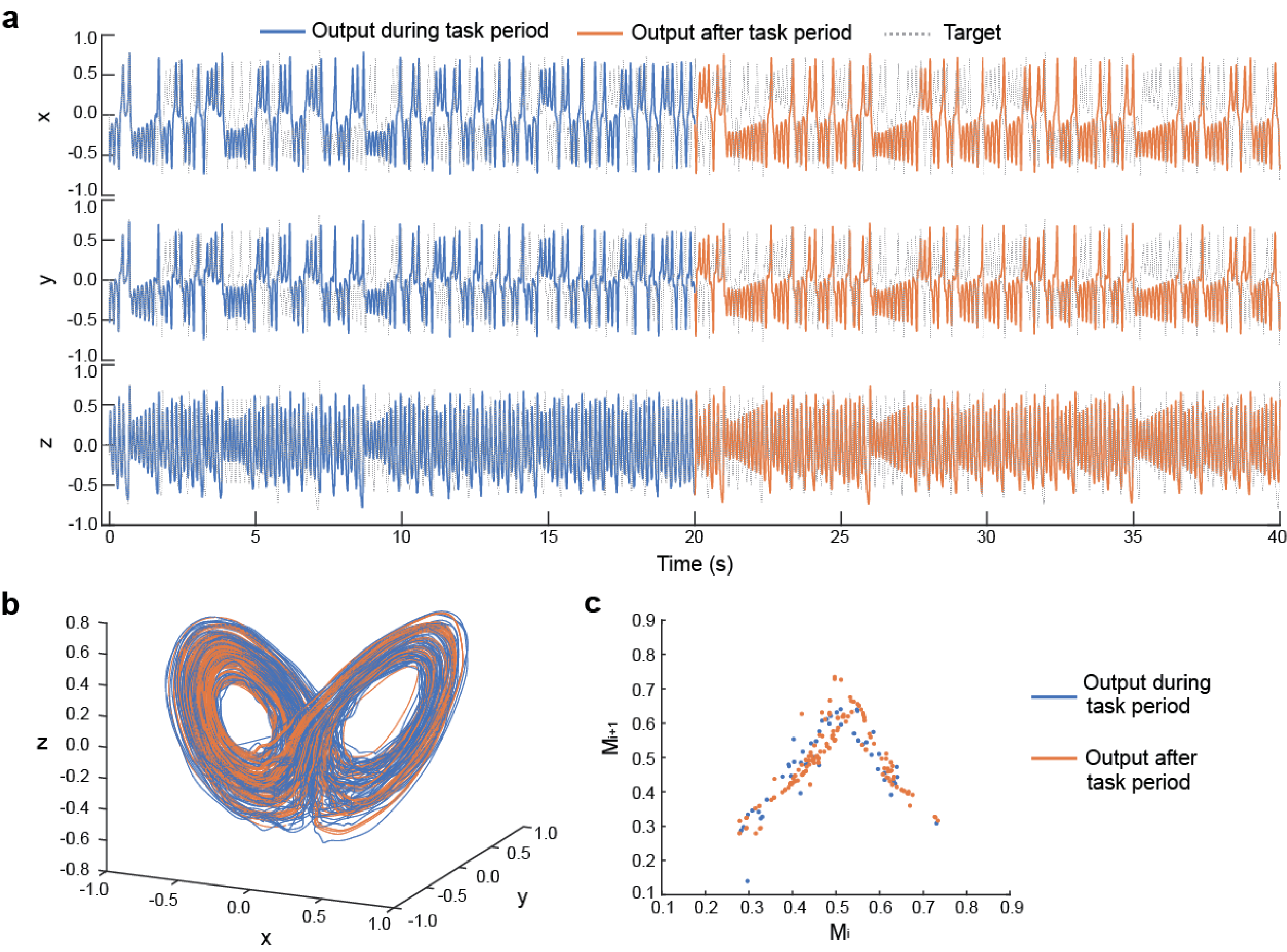}
		\caption{Example result of sinusoidal oscillation-driven reservoir computing with $[25, 50]$ Hz for the Lorenz time-series prediction task.
			{\bf a} Three-dimensional output for 40 s after training for 20 s. The blue and orange curves indicate output during and after the task period of 20 s, respectively.
			The broken gray curve indicates the target Lorenz time series.
			{\bf b} Three-dimensional plot of the output trajectory.
			{\bf c} Return map of successive maxima of the output for $z$-coordinate.}
		\label{fig:sup_25-50}
	\end{center}
\end{figure}

Example results of the sine ODRC using $[1, 10]$ and $[25, 50]$ Hz are shown in Figs.~\ref{fig:sup_5-10} and \ref{fig:sup_25-50}, respectively.
The ODRC output using low frequencies ($[1, 10]$ Hz) successfully reproduced the target during the task period, whereas the $z$-coordinate return map after the task period did not have a tent shape, suggesting poor generalization.
Nonetheless, the return map of the ODRC output using high frequencies ($[25, 50]$ Hz) had a tent shape, suggesting good generalization; however, it failed to reproduce the target during the task period.

\subsection{Results of neural ODRC in chaotic time-series prediction}
\label{appendix5}
\begin{figure}[!tbhp]
	\begin{center}
		\includegraphics*[width=13.0cm]{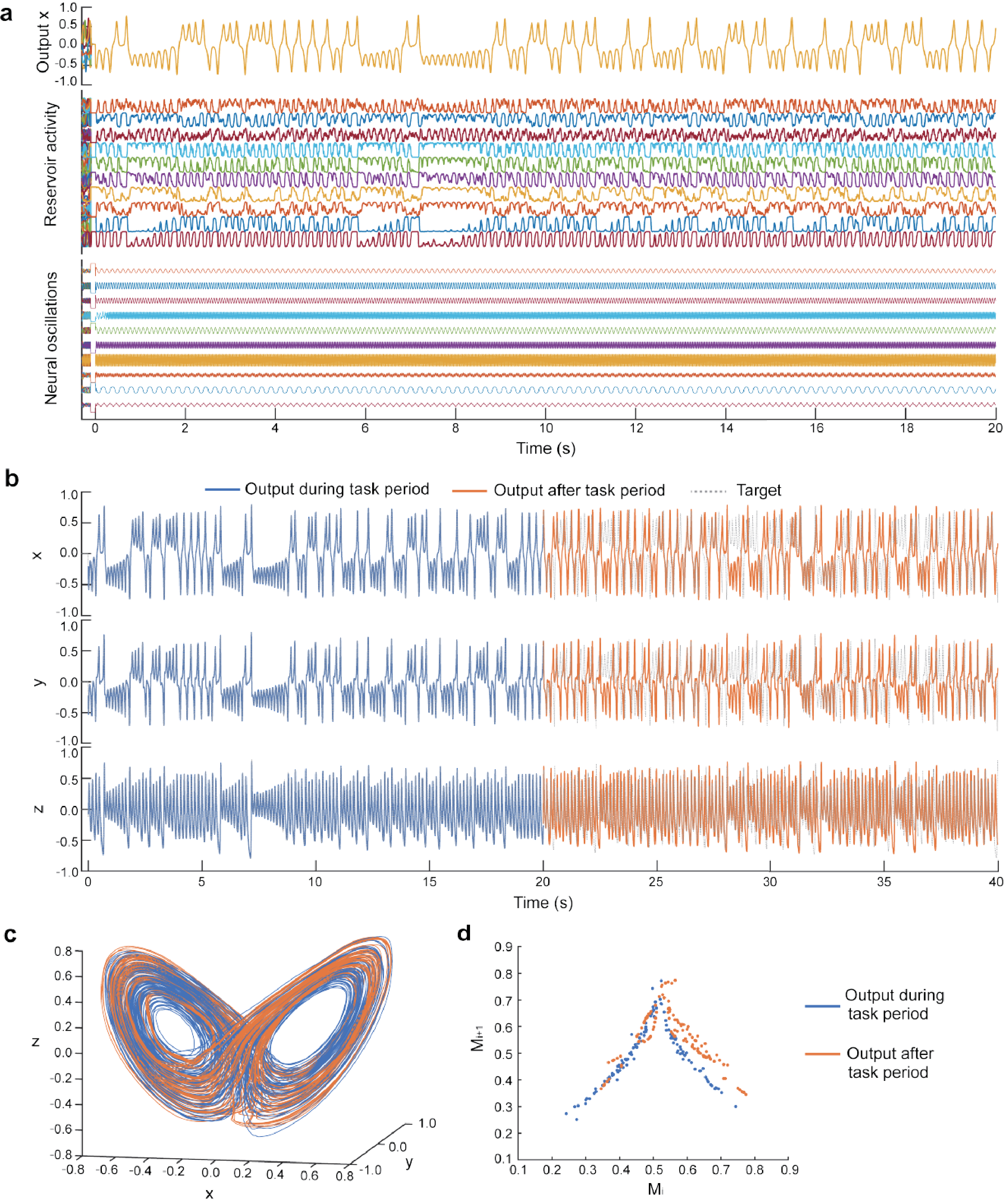}
		\caption{Example result of neural oscillation-driven reservoir computing for the Lorenz time-series prediction task.
			The time constant of the oscillator networks was 2 ms.
			{\bf a} Trajectories of the output for $x$-coordinate (top), reservoir activity of ten neural units (middle), and oscillations (bottom) in the 20-s task.
			These are the overlaid curves from ten test trials.
			{\bf b} Three-dimensional output for 40 s after training for 20 s.
			The blue and orange curves indicate the outputs during and after a task period of 20 s, respectively.
			The broken gray curve indicates the target Lorenz time series.
			{\bf c} Three-dimensional plot of the output trajectory.
			{\bf d} Return map of successive maxima of the output for $z$-coordinate.}
		\label{fig:sup_neural_lorenz}
	\end{center}
\end{figure}

\begin{figure}[!tbhp]
	\begin{center}
		\includegraphics*[width=13.0cm]{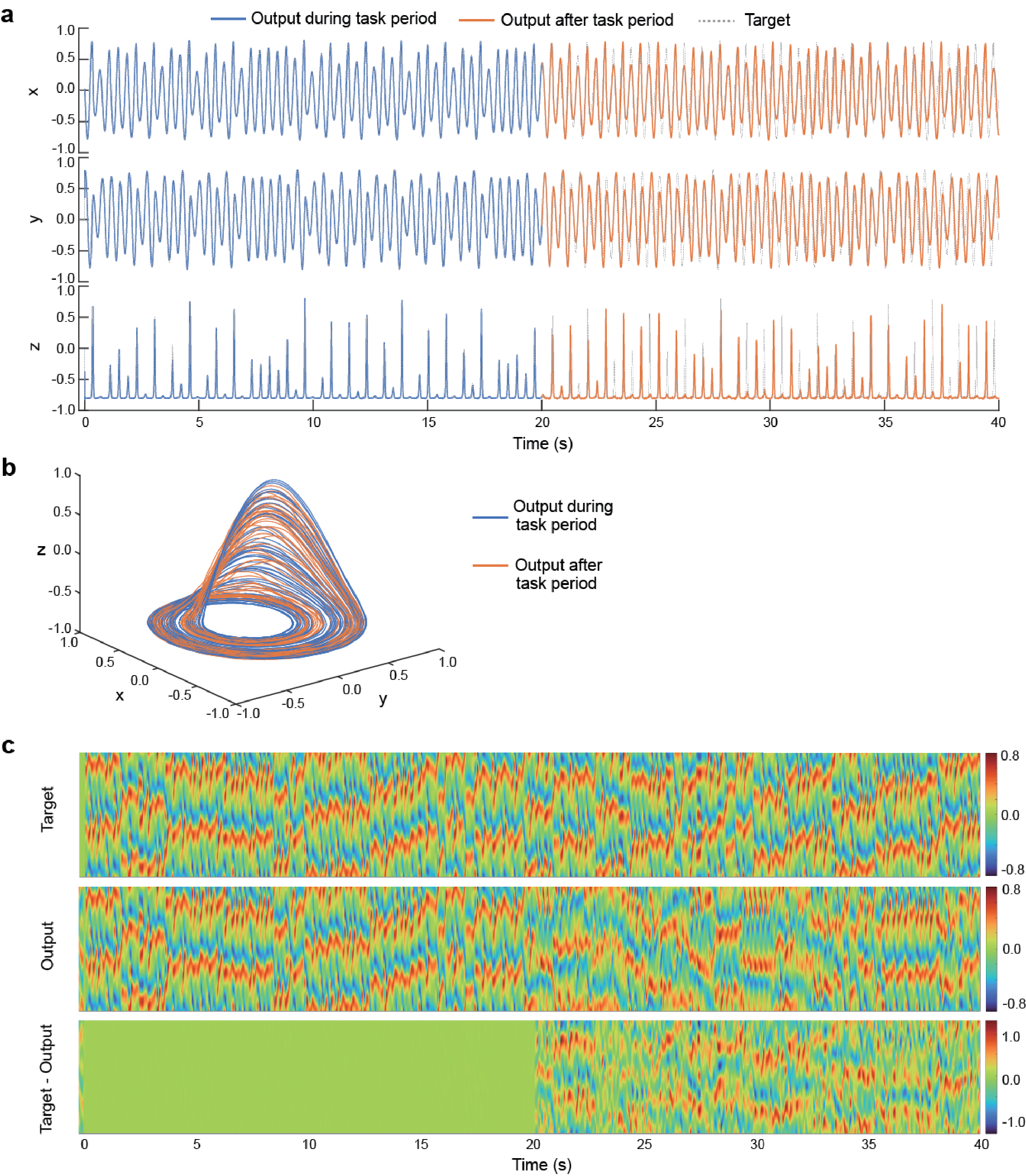}
		\caption{Example result of neural oscillation-driven reservoir computing (ODRC) for the R\"{o}ssler and Kuramoto--Sivashinsky (KS) time-series prediction tasks (20 s).
			{\bf a} Three-dimensional output for 40 s after training the R\"{o}ssler time series for 20 s.
			The blue and orange curves indicate the outputs during and after a task period of 20 s, respectively.
			The broken gray curve indicates the target R\"{o}ssler time series.
			{\bf b} Three-dimensional plot of the output trajectory for the R\"{o}ssler time-series prediction task.
			{\bf c} Contour plots of the target KS time series (top), ODRC output (middle), and the prediction differences from the target for the output (bottom).}
		\label{fig:sup_neural_rossler+ks}
	\end{center}
\end{figure}

\begin{figure}[!tbhp]
	\begin{center}
		\includegraphics*[width=13.0cm]{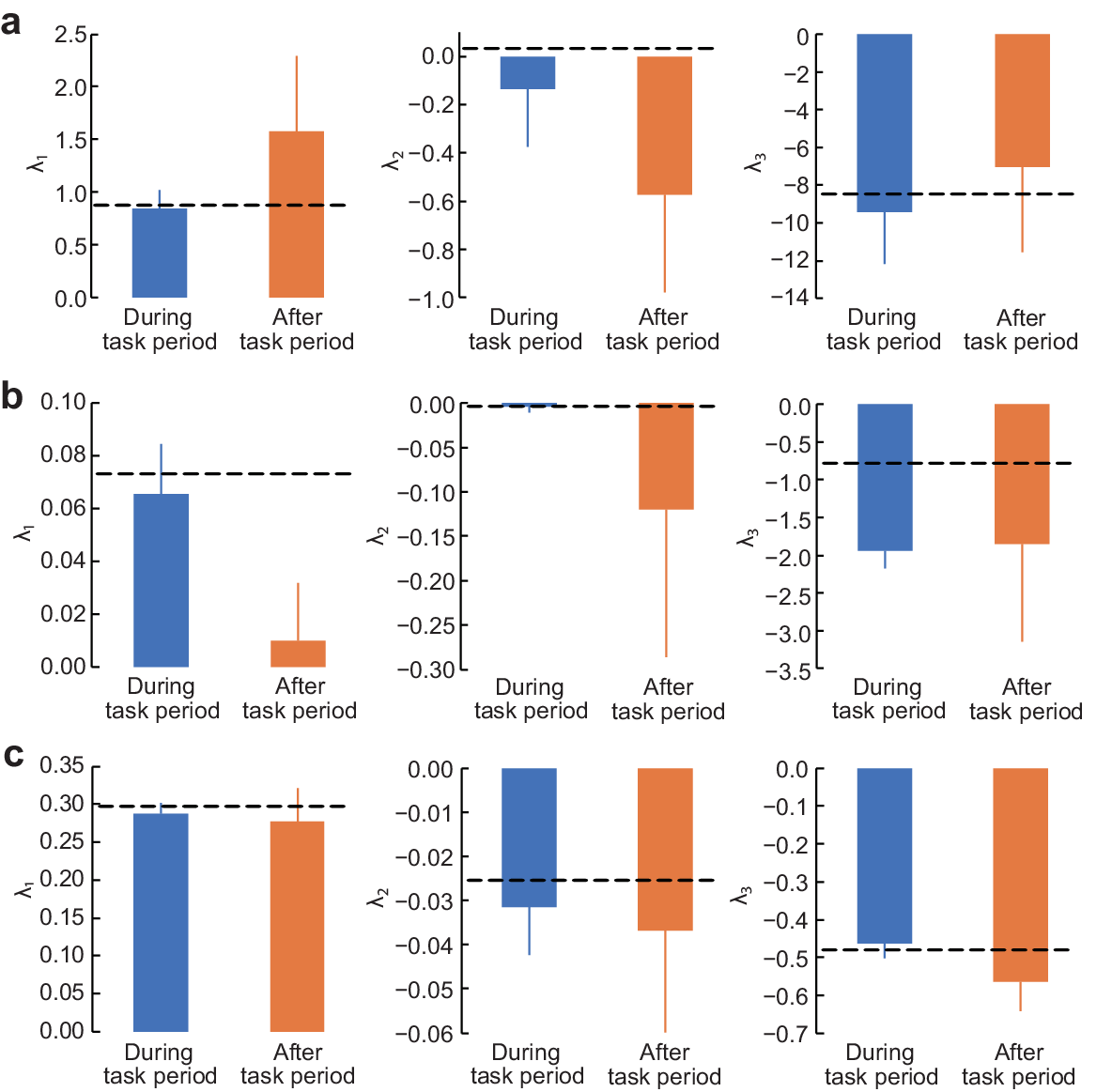}
		\caption{Estimated Lyapunov spectrum for neural oscillation-driven reservoir computing (ODRC).
			The horizontal broken lines indicate the Lyapunov exponents of the target.
			The blue and orange bars indicate the Lyapunov exponents of the outputs of the neural ODRC during and after the task period (20 s), respectively.
			The Lyapunov exponents were averaged over ten networks (mean $\pm$ s.d.).
			{\bf a}, {\bf b}, {\bf c} Lyapunov spectra for the Lorenz, R\"{o}ssler, and Kuramoto--Sivashinsky time-series predictions, respecitvely.}
		\label{fig:sup_neural_lyapunov_spectrum}
	\end{center}
\end{figure}

The results of the neural ODRC, where the time constant of the oscillator networks was 2 ms, were similar to those of the sine ODRC at $[10, 25]$ Hz.
Fig.~\ref{fig:sup_neural_lorenz} shows an example of the results of the neural ODRC in the Lorenz time-series prediction task.
The output can replicate and generalize the target time series.
Similar results were obtained for the R\"{o}ssler and KS time-series prediction tasks (Fig.~\ref{fig:sup_neural_rossler+ks}).
The results of the Lyapunov spectrum analysis support this conclusion (Fig.~\ref{fig:sup_neural_lyapunov_spectrum}).

\section*{Acknowledgements}
This work was supported by JST, CREST (JPMJCR17A4), PRESTO (JPMJPR23S5), ACT-X (JPMJAX21AN) and the project, JPNP16007, commissioned by the New Energy and Industrial Technology Development Organization (NEDO).

\bibliographystyle{elsarticle-harv}
\bibliography{references}






\end{document}